%% file: main.tex
\documentclass[10pt,twocolumn,letterpaper]{article}

\usepackage{cvpr}              


\usepackage[noend]{algpseudocode}
\usepackage{algorithmicx}
\usepackage{caption}
\usepackage{subcaption}
\usepackage{tabularx}
\usepackage{multirow}
\usepackage{pifont}
\usepackage{graphicx}
\usepackage{subfiles}
\usepackage{booktabs}
\usepackage{amsmath}
\usepackage{amssymb}
\DeclareMathOperator*{\argmax}{arg\,max}
\usepackage{algorithm}
\usepackage{adjustbox}
\usepackage{tabularx}
\usepackage{enumitem}
\usepackage[accsupp]{axessibility}
\usepackage{pifont}
\newcommand{\cmark}{\ding{51}}%
\newcommand{\xmark}{\ding{55}}%
\newcommand{\ours}{Cropper}

\definecolor{cvprblue}{rgb}{0.21,0.49,0.74}
\usepackage[pagebackref,breaklinks,colorlinks,allcolors=cvprblue]{hyperref}

\def\confName{CVPR}
\def\confYear{2025}

\title{\ours{}: Vision-Language Model for \\ Image Cropping through In-Context Learning}

\author{Seung Hyun Lee$^{2,4\ast\dagger}$, Jijun Jiang$^{3\ast}$, Yiran Xu$^{1,5\ast\dagger}$, Zhuofang Li$^{3\ast}$, Junjie Ke$^{1}$, Yinxiao Li$^{1}$, Junfeng He$^{2}$, \\ Steven Hickson$^{1}$, Katie Datsenko$^{1}$, Sangpil Kim$^{6}$, Ming-Hsuan Yang$^{1}$, Irfan Essa$^{1}$, Feng Yang$^{1\ddagger}$ \\ 
Google DeepMind$^{1}$, Google Research$^{2}$, Google$^{3}$, \\ University of Michigan$^{4}$, University of Maryland$^{5}$, Korea University$^{6}$ }

\begin{document}
\twocolumn[{
\renewcommand\twocolumn[1][]{#1}
\maketitle
\begin{center}
    \centering
    \captionsetup{type=figure}
    \includegraphics[width=0.85\linewidth]{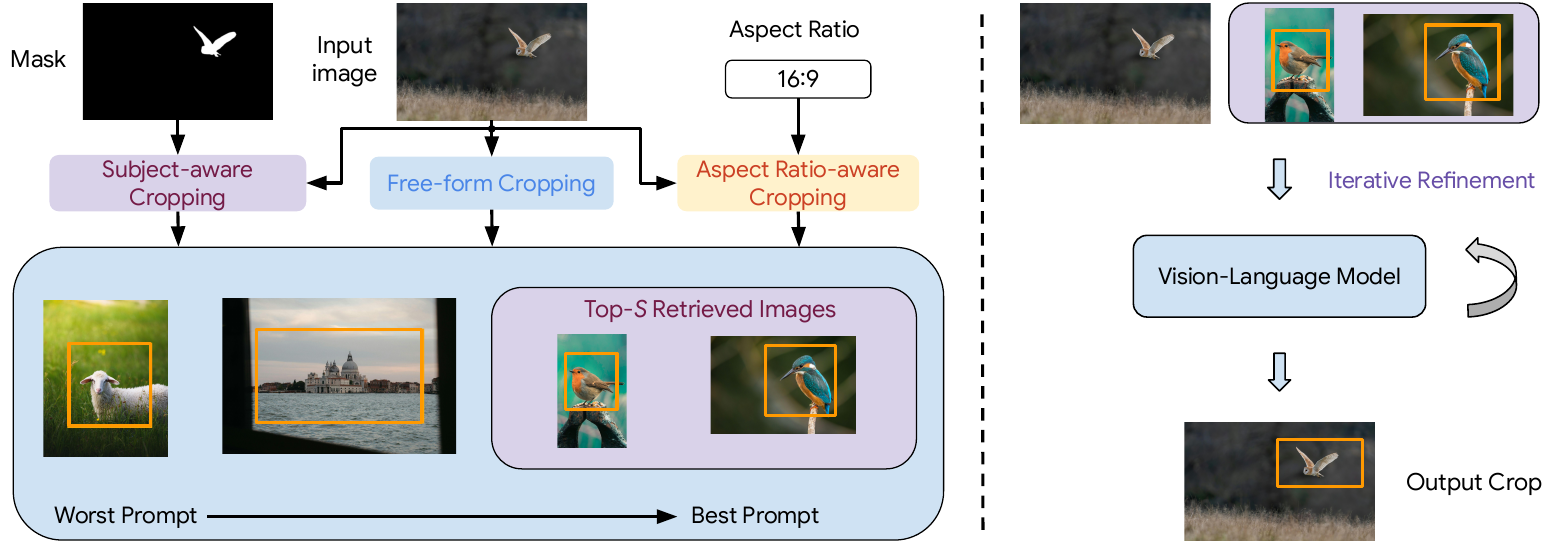}
    \vspace{-0.5em}
    \captionof{figure}{\textbf{\ours{}} is a unified framework for various cropping tasks, including free-form cropping, subject-aware cropping, and aspect ratio-aware cropping built on top of a pretrained large vision-language model through in-context learning. Given the input image, top-K semantically similar images are retrieved as in-context learning prompt, and fed to pretrained vision-language model to generate crops. The crop candidates are iteratively refined to yield the visually pleasing output crop. All images are from Unsplash~\cite{unsplash2025}.}
    \label{fig:fig0}
    \vspace{-0.2em}
\end{center}%
}]

\let\thefootnote\relax\par\footnote{\scriptsize{$\ast$ Equal contribution. $\dagger$ Work done while working at Google. $\ddagger$ Project lead.}}

\subfile{sec/0-abstract}

\subfile{sec/1-introduction}

\subfile{sec/2-related}    
\subfile{sec/3-method}    
\subfile{sec/4-result}

\subfile{sec/6-conclusion}
{
    \small
    \bibliographystyle{ieeenat_fullname}
    \bibliography{main.bib}

}
\newpage
\input{supplementary_arxiv}

\end{document}

%% file: sec/0-abstract.tex
\vspace{-1.0em}
\begin{abstract}

\vspace{-1.0em}
The goal of image cropping is to identify visually appealing crops in an image. Conventional methods are trained on specific datasets and fail to adapt to new requirements. Recent breakthroughs in large vision-language models (VLMs) enable visual in-context learning without explicit training. However, downstream tasks with VLMs remain under explored. In this paper, we propose an effective approach to leverage VLMs for image cropping. First, we propose an efficient prompt retrieval mechanism for image cropping to automate the selection of in-context examples. Second, we introduce an iterative refinement strategy to iteratively enhance the predicted crops. The proposed framework, we refer to as  Cropper, is applicable to a wide range of cropping tasks, including free-form cropping, subject-aware cropping, and aspect ratio-aware cropping. Extensive experiments demonstrate that Cropper significantly outperforms state-of-the-art methods across several benchmarks.
\end{abstract}

%% file: sec/1-introduction.tex
\section{Introduction}
\vspace{-0.75em}
\label{sec:intro}
Existing cropping methods~\citep{chen2016automatic,chen2017quantitative,chen2017learning,cheng2010learning,cheng2022re,li2018a2,li2020composing,pan2021transview,sun2013scale,tu2020image,wei2018good,yan2013learning,zhang2013weakly,zhang2005auto,zhong2021aesthetic} train neural networks on images and ground-truth crops to localize aesthetic crops. However, these approaches often depend on specially designed networks or features, which struggle to generalize effectively when confronted with new requirements or diverse datasets. Additionally, for specialized cropping tasks such as subject-aware cropping with subject masks or aspect ratio-aware cropping with target aspect ratio, unique networks must be developed and retrained, further complicating the process. This limitation underscores the need for more generalizable and versatile techniques in the field of image cropping.

Recent advancements in large vision-language models (VLM), such as GPT-4o~\citep{achiam2023gpt} and Gemini~\citep{team2023gemini}, have unlocked new potential for various vision tasks. Unfortunately, in a lot of cases, users are not able to fine-tune the VLM for downstream tasks. Effectively adapting large blackbox models for downstream tasks is very difficult. Luckily, in-context learning (ICL) ability is observed in large models~\citep{radford2021learning}. 
Given a test instance and a few in-context example demonstrations as input, the model directly infers the output without any parameter update or explicit training for the unseen task. ICL originates from natural language processing (NLP), and it has only recently been explored in the vision realm, mainly in image-to-image tasks~\citep{zhang2024makes,wang2023images,zhang2024instruct,bar2022visual}. In this paper, we undertake an investigation aimed at harnessing the power of VLMs through ICL for image cropping, which, to our knowledge, has not been explored before.

Despite the remarkable capabilities of VLMs, challenges persist. First, the effectiveness of visual ICL heavily relies on the quality of the in-context examples (i.e.  prompts)\citep{zhang2024makes,zhang2024instruct}. Manual selection of these examples would be laborious and difficult to scale. Moreover, how to incorporate aesthetics in VLM for image cropping is not straightforward. Leveraging VLM in-context learning for image cropping presents a novel research area requiring effective strategies.

To address these challenges, we propose an effective framework to adapt VLM for image cropping through in-context learning, referred to as \ours{}. It not only addresses the inherent challenges in traditional image cropping methods but also demonstrates versatility across various cropping tasks, including free-form cropping, subject-aware cropping, and aspect ratio-aware cropping. Illustrated in Fig.~\ref{fig:fig0}, our approach begins with an efficient prompt retrieval mechanism for image cropping tasks, automating the selection of relevant in-context examples to enhance efficiency. To further improve the performance, we introduce an iterative refinement strategy designed to enhance the quality of the predicted crops produced by VLM. To validate the efficacy of \ours{}, we conduct extensive experiments on various benchmark datasets.
\ours{} significantly outperforms existing state-of-the-art methods across various performance metrics. Notably, with only a few in-context examples, \ours{} achieves superior performance without the need for training. It also provides a unified framework for various cropping tasks. Our contributions are:

\begin{itemize}
    \item We introduce a unified visual in-context-learning framework \ours{} for image cropping tasks, including free-form, subject-aware, and aspect ratio-aware cropping.
    \item Our prompt retrieval strategy automates the effective selection of ICL examples for cropping tasks.
    \item The proposed iterative refinement strategy enables the model to progressively enhance the output crop.
    \item With a few in-context examples and no explicit training, \ours{} surpasses the existing supervised learning methods across various benchmarks.
\end{itemize}

%% file: sec/2-related.tex
\section{Related Work}
\label{sec:related}
\vspace{-0.5em}

\begin{figure*}[t!]
  \centering
  \includegraphics[width=0.9\linewidth]{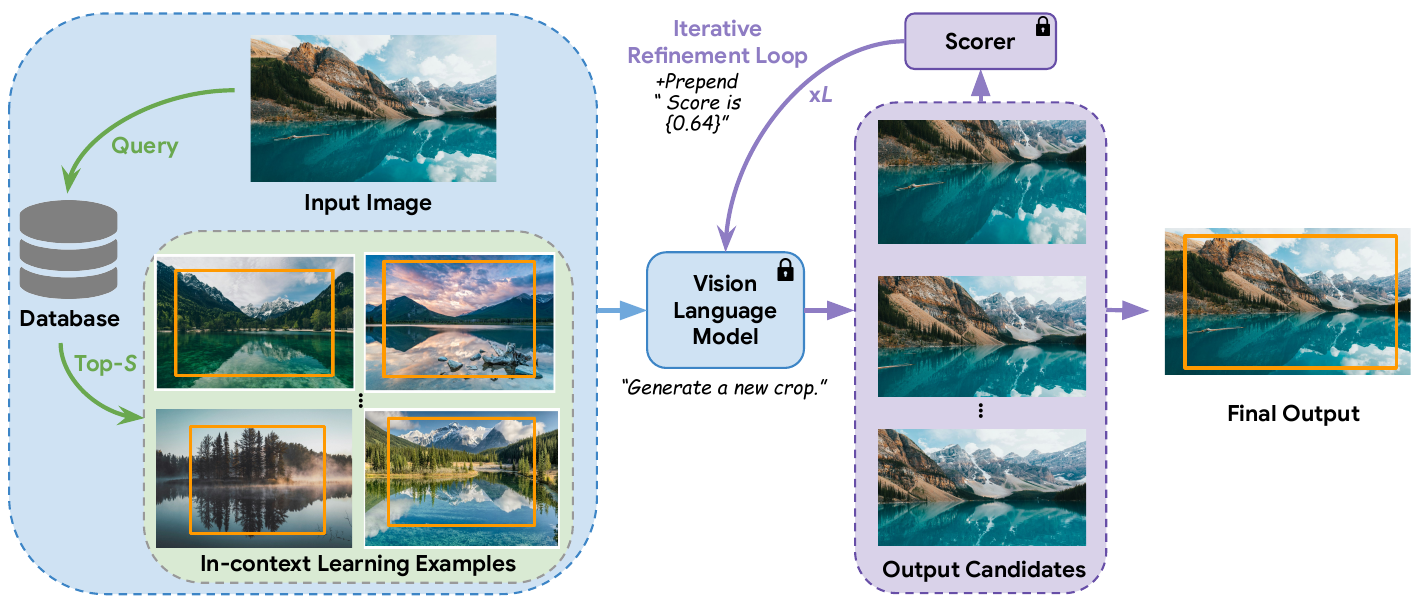}
  \caption{\textbf{\ours{} Overview}. \ours{} consists of two main steps: visual prompt retrieval and iterative crop refinement. Through visual prompt retrieval, top-$S$ ICL examples are retrieved using an image similarity metric. In the iterative crop refinement stage, the VLM generates candidate crops based on these ICL examples and then these crops are subsequently scored by a scorer which measures aesthetics, content similarity, and area size. The VLM iteratively refines the crop candidates using the feedback from the scorer $L$ times. All images are from Unsplash~\cite{unsplash2025}.}
  \vspace{-1em}
  \label{fig:overview}
\end{figure*}

\noindent \textbf{Image Cropping.}
Image cropping is a critical operation for various photography-related applications. From the perspective of constraints, there are three commonly studied types of cropping problems. The first category is free-form cropping, where the objective is to directly identify the best crop without imposing additional constraints. Numerous techniques have been explored to tackle this problem, including saliency maps~\citep{tu2020image}, learning-based methods~\citep{chen2017quantitative,chen2017learning,cheng2022re,deng2018aesthetic,guo2018automatic,hong2021composing,kao2017automatic,li2020composing,lu2019listwise,mai2016composition,tu2020image,wei2018good,zeng2019reliable,zeng2020grid}, and reinforcement learning~\citep{li2018a2}. Another cropping task is subject-aware image cropping~\citep{hong2023learning,yang2023focusing}, where an additional subject mask is provided to indicate the subject of interest. The third task is aspect ratio-aware cropping~\citep{9156674}, where the crops are expected to adhere to a specified aspect ratio. Most existing image cropping approaches rely on training neural networks on specific datasets, requiring retraining to accommodate different data distributions and requirements. In contrast, our method requires only a few in-context examples and doesn't need explicit training. Moreover, none of the methods are flexible enough to handle all three cropping tasks in unified manner, while our method can do.

\noindent \textbf{In-Context Learning.}
In-context learning is a recent paradigm originating from NLP, where large-scale models perform inference on unseen tasks by conditioning on a few in-context examples and the test instance.  This paradigm is effective because users can directly adapt the model to different downstream tasks without the hassle of fine-tuning or changing the model parameters in any way. Numerous methods based on ICL have been developed for various tasks such as text classification~\citep{zhang-etal-2022-active} and machine translation~\citep{zhang2023prompting}. ICL is still relatively new in computer vision, and most visual ICL works focus on using large-scale image-to-image vision models for tasks such as image inpainting~\citep{bar2022visual} and segmentation~\citep{wang2023images,zhang2024instruct}. Empowered by recent breakthroughs in VLMs such as OpenAI GPT-4o~\citep{achiam2023gpt} and Google Gemini~\citep{team2023gemini}, we investigate effective strategies for in-context learning for cropping for the first time.

\noindent \textbf{Prompt Retrieval.}
NLP researchers have discovered that the selection and arrangement of in-context examples, also known as prompts, significantly impacts the output performance~\citep{lu2021fantastically,zhang2024makes}. These findings have sparked interest in prompt retrieval, where in-context learning examples are retrieved based on similarity metrics given a test instance. Liu~et al.~\citep{zhang2024makes} have demonstrated success in selecting semantically similar in-context examples based on nearest neighbors measured by embeddings from a pretrained languages model. Rubin~et al.~\citep{rubin2021learning} propose selecting examples using a supervised prompt retriever to maximize downstream performance. For visual ICL for image-to-image tasks, Zhang et al.~\citep{zhang2024makes} use CLIP~\citep{radford2021learning}-based unsupervised embedding similarity measure, and demonstrate further improvement with a supervised prompt retrieval approach.

%% file: sec/3-method.tex
\section{Method}
\vspace{-0.5em}
\label{sec:method}
Fig.~\ref{fig:overview} illustrates the structure of \ours{}, which consists of two main steps: visual prompt retrieval on the left and feedback-based iterative refinement on the right. Given the input image, \ours{} automatically retrieves the top-$S$ suitable in-context learning examples along with their ground-truth crop coordinates. Both the input image and the retrieved in-context learning examples are then fed into the vision-language model. The model is prompted to propose several potential crop candidates represented by their coordinates. In the iterative refinement stage, crops are generated based on the output of the VLM. These candidate crops are then evaluated using an  aesthetic scorer, CLIP~\cite{radford2021learning} scorer, and area size, which provides feedback guidance for the VLM. The model then iteratively refines the crop candidates based on this feedback. The iterative refinement process repeats $L$ times to produce the final output.

\vspace{-0.5em}
\subsection{Visual Prompt Retrieval for Cropping}
\vspace{-0.5em}
\label{sec:retrieval}
The simplest method for retrieving ICL examples is random selection, where one or multiple samples are randomly chosen from the training dataset. However, previous studies have demonstrated that the ICL performance of such random selection is highly sensitive to the chosen samples~\citep{lu2021fantastically,zhang2024makes}. In our experiments (Sec.~\ref{sec:result}), we empirically confirm that random prompt selection often leads to suboptimal results. Therefore, our objective is to explore an effective strategy for automatically selecting the most suitable examples for various cropping tasks.

Intuitively, similar images are more likely to be cropped similarly. Thus, we aim to retrieve the top-$S$ images and their most relevant ground-truth crops based on some similarity metric. Formally, given an image query $z_q$ and a dataset $\mathcal{D}={(z_i,C_i)}^{M}_{i=1}$ containing $M$ pairs of image $z_i$ and crop ground-truth $C_i$, where $C_i$ contains multiple crops $c_1, \dots, c_s$ for some datasets, we seek to retrieve the most relevant in-context examples and crop ground-truth as:
\setlength{\abovedisplayskip}{4pt} 
\setlength{\belowdisplayskip}{4pt} 
\begin{align}
\mathcal{Z} &= \argmax_{ z_i \in \mathcal{D}}  Q(z_q, z_i),\ \ |\mathcal{Z}| = S, \label{eq:image_similarity} \\
\mathcal{H} &= \argmax_{ c_j \in C_j} G(z_q, c_j),\ \ z_j \in \mathcal{Z},\ \ |\mathcal{H}| = T,
\label{eq:crop_similarity}
\end{align}
where $\mathcal{Z}$ represents the set of top-$S$ relevant images selected from the dataset $\mathcal{D}$ based on the similarity metric $Q(z_q, z_i)$. $\mathcal{H} = (z_j, c_j)_{j=1}^S$ represents the selected in-context images $z_j$ along with their most relevant $T$ crop ground-truths based on metric $G(z_q, c_j)$.  $Q$ and $G$ are designed differently to accommodate different cropping tasks, including free-form cropping, subject-aware cropping, and aspect ratio-aware cropping.

\begin{table*}[tp]
    \centering
    \scriptsize
    \begin{tabularx}{\linewidth}{lX}
        \toprule
        \textbf{Prompt \& Output} & \textbf{Instruction}  \\
        \midrule
        Initial Prompt & ``Localize the aesthetic part of the image. $(s,x_1,y_1,x_2,y_2)$ represents the region. $x_1$ and $x_2$ are the left and right most positions, normalized into 1 to 1000, where 1 is the left and 1000 is the right. $y_1$ and $y_2$ are the top and bottom positions, normalized into 1 to 1000 where 1 is the top and 1000 is the bottom. $s$ is MOS score. We provide several images here.  \\
        & \{image 1\} $(s_1^{1},x^{1,1}_1,y^{1,1}_1,x^{1,1}_2,y^{1,1}_2)$, $(s_1^{2},x_1^{1,2},y_1^{1,2},x_2^{1,2},y_2^{1,2})$, ..., $(s_1^{T},x_1^{1,T},y_1^{1,T},x_2^{1,T},y_2^{1,T})$,  \\
        & \{image 2\}, $(s_2^{1},x_1^{2,1},y_1^{2,1},x_2^{2,1},y_2^{2,1})$, $(s_2^{2},x_1^{2,2},y_1^{2,2},x_2^{2,2},y_2^{2,2})$, ..., $(s_2^{T},x_1^{2,T},y_1^{2,T},x_2^{2,T},y_2^{2,T})$,\\
        & ... \\ 
        & \{image S\}, $(s_{S}^{1},x_1^{S,1},y_1^{S,1},x_2^{S,1},y_2^{S,1})$, $(s_{S}^{2},x_1^{S,2},y_1^{S,2},x_2^{S,2},y_2^{S,2})$, ..., $(s_{S}^{T},x_1^{S,T},y_1^{S,T},x_2^{S,T},y_2^{S,T})$,\\
        & \{Query image\}, \\
        \textbf{Output} & $(\hat{s}_1,\hat{x}_1^1,\hat{y}_1^1,\hat{x}_2^1,y_2^1)$,$(\hat{s}_2,\hat{x}_1^2,\hat{y}_1^2,\hat{x}_2^2,\hat{y}_2^2)$, ..., $(\hat{s}_R,\hat{x}_1^R,\hat{y}_1^R,\hat{x}_2^R,\hat{y}_2^R)$  \\
         \midrule
        Iterative Crop  &  Initial Prompt + \{Cropped image 1\} $(\hat{s}_1,\hat{x}_1^1,\hat{y}_1^1,\hat{x}_2^1,\hat{y}_2^1)$, Score is \{score 1\} \\
        Refinement Prompt & \{Cropped image 2\} $(\hat{s}_2,\hat{x}_1^2,\hat{y}_1^2,\hat{x}_2^2,\hat{y}_2^2)$, Score is \{score 2\} \\
        & ... \\
         & \{Cropped image R\} $(\hat{s}_R,\hat{x}_1^R,\hat{y}_1^R,\hat{x}_2^R,\hat{y}_2^R)$, Score is \{score R\} \\
          & Propose similar crop that has high score. The region should be represented by $(s,x_1,y_1,x_2,y_2)$. \\ 
          \textbf{Output} & $(\hat{\hat{s}},\hat{\hat{x}}_1,\hat{\hat{y}}_1,\hat{\hat{x}}_2,\hat{\hat{y}}_2)$ \\
        \bottomrule
    \end{tabularx}
    \caption{VLM prompt used for free-form cropping. The goal is to find the most visual pleasing crop $(\hat{\hat{s}},\hat{\hat{x}}_1,\hat{\hat{y}}_1,\hat{\hat{x}}_2,\hat{\hat{y}}_2)$. In the initial prompt, we use $S$ in-context~(ICL) examples, and $T$ ground-truth crops. The format of image $i$'s $j$-th crop is defined as $(s_i^j, x_1^{i, j}, y_1^{i, j}, x_2^{i, j}, y_2^{i, j})$. Intermediate results of initial prompt are coordinates of $R$ crops. Subsequently, the crop is iteratively refined by accumulating the context into prompts, using refinement prompt. Note that $\{\text{score}\}$ is calculated based on VILA~\cite{ke2023vila} and area size. }
    \label{tab:free_main} 
    \vspace{-1em}
\end{table*}

\noindent \textbf{Free-form cropping} aims to identify the best crop without additional constraints regarding aspect ratio or target subject. For this cropping task, we use the~CLIP~\citep{radford2021learning} image embeddings as an off-the-shelf image feature extractor, and $Q$ corresponds to the cosine similarity between the input image $z_q$ and each training example $z_i \in \mathcal{D}$. In free-form cropping datasets, such as GAICD~\citep{zeng2020grid}, each image $z_i$ is associated with multiple ground-truth crops $c_i$, each with its mean opinion score (MOS) aggregated from human evaluation. We use the MOS score as $G$ for selecting the ground-truth crops. Therefore, after obtaining $\mathcal{Z}$, we select the top-ranked crops  based on their MOS. Each crop ground-truth $c_i$ is represented as a 5-tuple, $(s,x_1,y_1,x_2,y_2)$, indicating the MOS and the leftmost, top, rightmost, and bottom positions, respectively.

\noindent \textbf{Subject-aware cropping } intends to identify an aesthetic crop containing the subject of interest, which is represented by binary masks provided by users. In this task, the query image $z_q$ is accompanied by a binary mask $m_q$ indicating the subject of interest. Similarly, we first use CLIP image embedding similarity as $Q$ for retrieving the top-$S$ relevant images. Since each image in this task is associated with multiple target subject masks and their corresponding ground-truth crops, we further refine it by choosing the most similar mask areas to provide better guidance. $G$ is defined as $-L_2$ distance between the center points of the target mask $m_q$ and mask from image $z\in \mathcal{Z}$ to select the crop with closest masks. As a result the ground-truth crop for the closest mask is provided as the in-context learning example label, and each crop ground-truth $c_i$ is represented  $(x_1,y_1,x_2,y_2)$.

\noindent \textbf{Aspect ratio-aware cropping} requires the generated crop to conform to a specified aspect ratio $r_q$  given the query image $z_q$. Each image in the dataset for this task is associated with ground-truth crops using different aspect ratios, such as 16:9, 3:4, and 1:1. Similarly, CLIP-based image similarity is adopted as $Q$. $G$ is defined as the similarity between the crop $c_i$'s aspect ratio and the target aspect ratio $r_q$. In other words, for each image $z\in \mathcal{Z}$, only the crop that has the similar target aspect ratio is used as in-context learning ground-truth. Similar to previous tasks, each crop ground-truth $c_i$ is represented  $(x_1,y_1,x_2,y_2)$. 
\vspace{-0.5em}
\subsection{Iterative Crop Refinement}
\label{sec:feedback}
\vspace{-0.5em}
Without explicit supervision, VLM lacks a deep understanding of the context of the cropping task, such as the provided coordinate system and intended aesthetics. Consequently, it often produces nonsensical outputs even when provided with good in-context learning cropping examples. Empirically, we observe that the initial crop candidates generated by the VLM lack diversity and sometimes fail to make sense (e.g. being too small or too large). Yang et al.~\citep{yang2024large} have shown that large language models can optimize the output by iteratively incorporating feedback. Motivated by this, we propose an iterative crop refinement mechanism to further guide the VLM in generating high quality crops. Concretely, the VLM is prompted to generate $R$ crop candidates based on the in-context learning examples retrieved using the method described in Sec.~\ref{sec:retrieval}. Subsequently, we crop the image according to each cropping proposal and feed the cropped images into scorers, such as VILA~\cite{ke2023vila} covering aesthetics, CLIP~\cite{radford2021learning} measuring content preserving, and area size, to obtain corresponding scores. In the refinement phase, we iteratively provide such feedback to the VLM by scoring the crop candidates and prompting it to generate new candidates to improve the score. This iterative process is repeated $L$ times to generate the final output. Tab.~\ref{tab:free_main} shows the prompt design for free-form cropping with the two phases of \ours{}. For subject-aware cropping and aspect ratio aware cropping, the only difference depends on whether MOS is predicted together.

%% file: sec/4-result.tex
\section{Experimental Results}
\vspace{-0.5em}
\label{sec:result}
We first describe the implementation details and experimental setups before 
presenting quantitative and qualitative results with comparisons to SOTA  as well as ablation studies.

\begin{figure*}[t!]
  \centering
  
  \begin{minipage}{0.32\textwidth}
    \centering
    \vspace{-0.8em}
    \includegraphics[width=\linewidth]{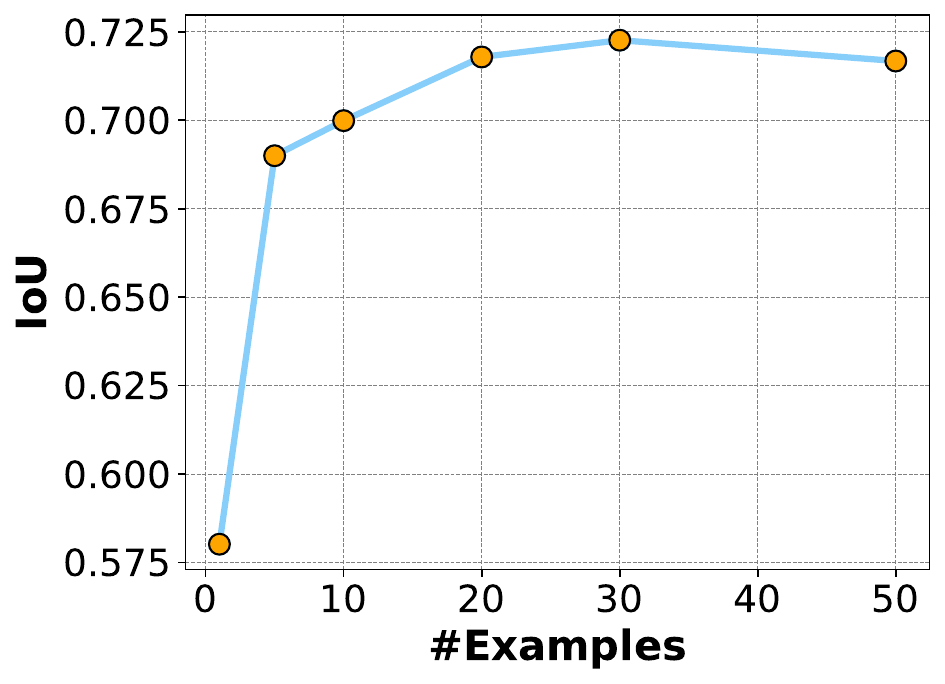}
    \vspace{-1.5em}
    \caption{Relationship between number of in-context learning examples $S$ and IoU on the GAICD~\citep{zeng2020grid} validation dataset for free-form cropping. We could see when the number of in-context learning examples $S$ is 30, IoU is the best.}
    \label{fig:hy_num_icl}
  \end{minipage}
  \hfill
  \begin{minipage}{0.32\textwidth}
    \centering
    \vspace{-3.0em}
    \includegraphics[width=\linewidth]{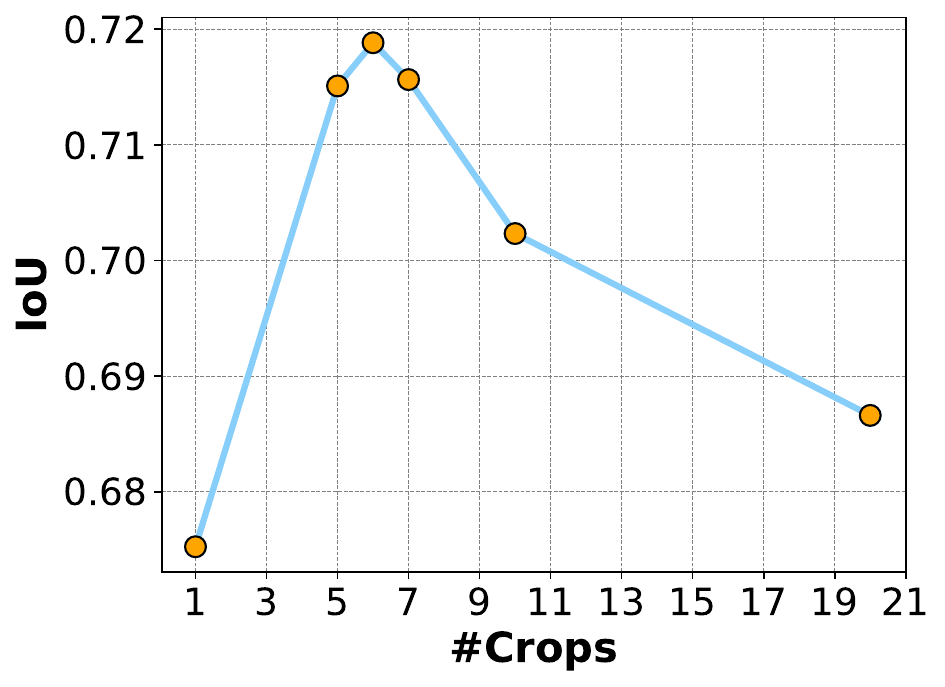}
    \vspace{-1.5em}
    \caption{Relationship between number of crops $R$ and IoU on the GAICD~\citep{zeng2020grid} validation dataset for free-form cropping. When the number of crops $R$ is 6, IoU is the best.}
    \label{fig:hy_num_crop}
  \end{minipage}
  \hfill
  \begin{minipage}{0.32\textwidth}
    \centering
    \vspace{-1.0em}
    \includegraphics[width=\linewidth]{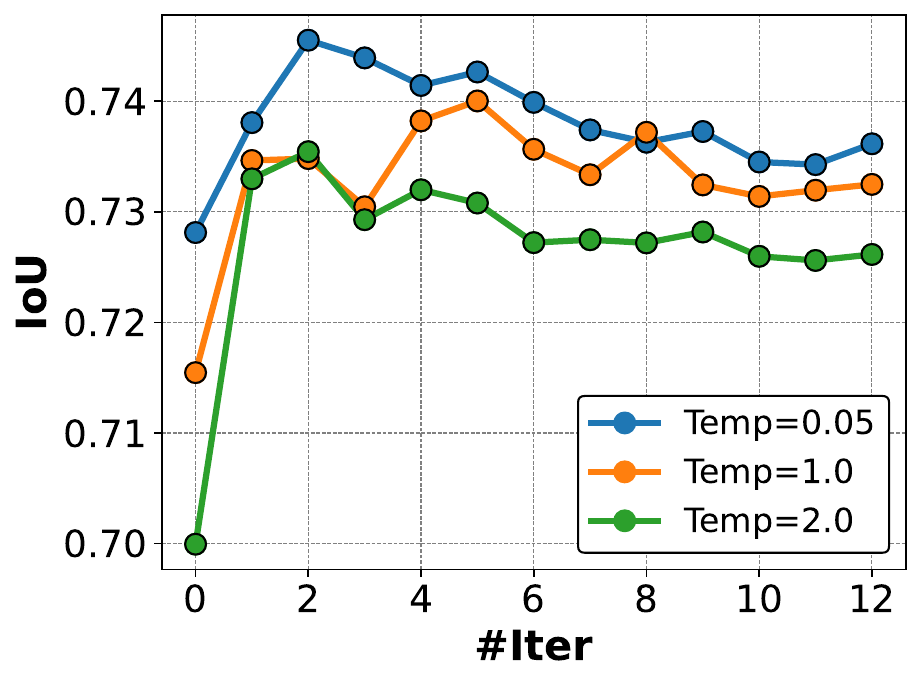}
    \vspace{-1.9em}
    \caption{Relationship among the number of refinement iterations $L$, VLM model temperature and IoU on the GAICD~\citep{zeng2020grid} validation dataset. The experiments are based on the optimal number of ICL examples and candidate crops previously determined.}
    \label{fig:number_feedback2}
  \end{minipage}
  \vspace{-1em}

\end{figure*}

\subsection{Implementation Details}
\label{subsec:detail}

\noindent \textbf{Dataset.} We evaluate \ours{} across three cropping benchmarks: GAICD~\citep{zeng2020grid} for free-form cropping, FCDB~\citep{chen2017quantitative} for free-form cropping and aspect ratio-aware cropping, and SACD~\citep{yang2023focusing} for subject-aware cropping. 

GAICD~\citep{zeng2020grid} dataset has 3,336 images, with 2,636 for training, 200 for validation, and 500 for testing, containing 288,069 densely annotated crops. For evaluation, we retrieve in-context learning examples from the GAICD~\citep{zeng2020grid} training set (the validation set is not used). Among the 90 annotations available for each retrieved image, we select the top-$T$ crops ranked by MOS. Additionally, employing the same retrieval strategy on the GAICD~\citep{zeng2020grid} training set, we further evaluate the performance on 348 test images from the FCDB~\citep{chen2017quantitative} dataset, measuring the out-of-domain performance. Following \citep{9156674}, we also use the FCDB~\citep{chen2017quantitative} dataset to evaluate the performance of aspect ratio-aware cropping, treating the aspect ratio of the user-annotated box as the expected aspect ratio. The subject-aware cropping dataset SACD~\citep{yang2023focusing} contains 2,906 images, with 2,326 for training, 290 for validation, and 290 for testing. 

\begin{table*}[h!]
  \centering
  \resizebox{\textwidth}{!}{\begin{tabular}{@{}lccccccccccccc@{}}
    \toprule
    Model  & $Acc_{1/5}$ & $Acc_{2/5}$ & $Acc_{3/5}$ & $Acc_{4/5}$ & $\overline{Acc}_{5}$ & $Acc_{1/10}$ & $Acc_{2/10}$ & $Acc_{3/10}$ & $Acc_{4/10}$ & $\overline{Acc}_{10}$ & $\overline{SRCC}$ & $\overline{PCC}$\\
    \midrule
    A2RL~\citep{li2018a2} & 23.2 & - & - & - & - & 39.5 & - & - & - & - & - & -  \\
    VPN~\citep{wei2018good} & 36.0 & -  & - & - & - & 48.5 & - & - & - & - & - & -  \\
    VFN~\citep{chen2017learning} & 26.6 & 26.5  & 26.7 & 25.7 & 26.4 & 40.6 & 40.2 & 40.3 & 39.3 & 40.1 & 0.485 & 0.503 \\
    VEN~\citep{wei2018good} & 37.5 & 35.0  & 35.3 & 34.2 & 35.5 & 50.5 & 49.2 & 48.4 & 46.4 & 48.6 & 0.616 & 0.662 \\
    GAIC~\citep{zeng2020grid} & 68.2 & 64.3 & 61.3 & 58.5 & 63.1 & 84.4 & 82.7 & 80.7 & 78.7 & 81.6 & 0.849 & 0.874 \\
    CGS~\citep{li2020composing} & 63.0 & 62.3  & 58.8 & 54.9 & 59.7 & 81.5 & 79.5 & 77.0 & 73.3 & 77.8 & 0.795 & -   \\
    TransView~\citep{pan2021transview} & 69.0 & 66.9  & 61.9 & 57.8 & 63.9 & 85.4 & 84.1 & 81.3 & 78.6 & 82.4 & 0.857 & 0.880   \\
    Chao et al.~\citep{wang2023image} & 70.0 & 66.9  & 62.5 & 59.8 & 64.8 & 86.8 & 84.5 & 82.9 & 79.8 & 83.3 & 0.872 & \textbf{0.893}   \\
    \midrule
    \ours{} (Ours) & \textbf{88.9} & \textbf{85.9} & \textbf{83.1} & \textbf{79.4}  & \textbf{84.3}  & \textbf{98.2}  & \textbf{97.2} & \textbf{96.4} & \textbf{94.3} & \textbf{96.5}  & \textbf{0.904}  & 0.860  \\
  \bottomrule
  \end{tabular}}
 \vspace{-0.5em}
  \caption{Quantitative comparison with existing free-form cropping methods on the GAICD~\citep{zeng2020grid} dataset. \ours{} demonstrates significant superiority over other baselines despite using only a few in-context learning examples and no explicit training.
  }
  \vspace{-0.5em}
  \label{tab:tab_crop}
\end{table*}

\noindent \textbf{Metrics.}
We use standard metrics in the image cropping community~\citep{zeng2020grid}, including the Spearman's rank-order correlation coefficient $\overline{SRCC}$, the Pearson correlation $\overline{PCC}$, and  $Acc_{K/N}$, which is also used to evaluate image cropping methods~\cite{li2018a2} generating arbitrary bounding boxes. These metrics quantify the alignment of the generated crops with aesthetic preferences, using the ground-truth mean opinion score~(MOS). Specifically, $\overline{PCC}$ assesses the linear correlation between the predicted MOS and the ground-truth MOS, whereas $\overline{SRCC}$ measures the correlation of ranking order. Given that \ours{} generates $R$ candidate crops per each iteration step on the GAICD~\citep{zeng2020grid} dataset, we compute $\overline{SRCC}$ and $\overline{PCC}$ using the best five crops instead of considering all crops. $Acc_{K/N}$ indicates whether top-$K$ from predictions could be involved among top-$N$ crops from the ground-truth based on MOS. $Acc_{1/5}$, $Acc_{2/5}$, $Acc_{3/5}$, $Acc_{4/5}$, $Acc_{1/10}$, $Acc_{2/10}$, $Acc_{3/10}$, $Acc_{4/10}$ are measured with $N \in \{5, 10\}$ and $K \in \{1,2,3,4\}$ to return $K$ of top-$N$ accuracy. Additionally, we use Intersection-over-Union (IoU) and Boundary-displacement-error (Disp) metrics to compare with other approaches on the FCDB~\citep{chen2017quantitative} and SACD~\citep{yang2023focusing} datasets. Disp represents the average $L_1$ distance between the ground-truth coordinates and the predicted values. To verify the effectiveness of the proposed method, we also conduct user study in Sec.~\ref{sec:user}.

\noindent \textbf{Vision-language model.}
We adopt publicly available Gemini 1.5 Pro~\cite{reid2024gemini} model via the Vertex AI API for our task, and experiment with GPT-4o~\citep{achiam2023gpt}. This is because they support visual prompting with many images, which is critical for in-context learning for image cropping.

\noindent \textbf{In-context learning examples.}
Similarity measurement $Q$ in Eq.~\ref{eq:image_similarity} is implemented using cosine similarity between image embeddings extracted from the ViT-B/32 variant of CLIP~\citep{radford2021learning}. 
As shown in Fig.~\ref{fig:hy_num_icl}, we studied the relationship between the number of in-context examples $S$ and IoU on the GAICD~\citep{zeng2020grid} validation dataset for free-form cropping and found the best number of in-context learning examples $S$ is 30. We set $S$ to 30 by default.

\noindent \textbf{Number of crops.}
To determine the number of crops $R$, as in Fig.~\ref{fig:hy_num_crop}, we studied the relationship between the number of crops and IoU on the GAICD~\citep{zeng2020grid} validation dataset for free-form cropping and found the best number of crops $R$ is 6. We set the number of crops $R$ to be 6 by default.

\begin{figure*}[t!]
  \centering
  \includegraphics[trim={0 0 0 0},clip,width=\linewidth]{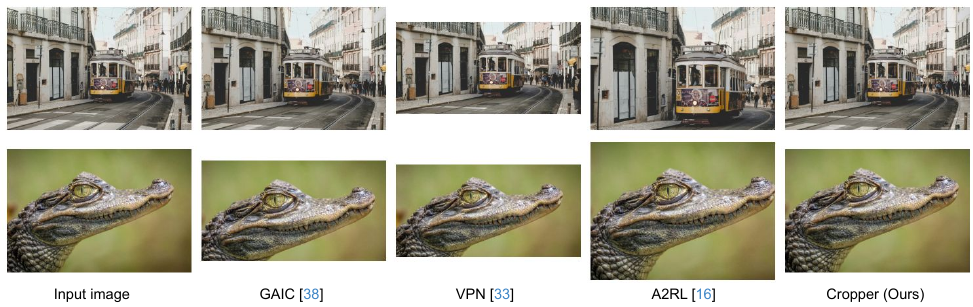}
   
  \caption{Qualitative comparing \ours{} with GAIC~\citep{zeng2020grid}, VPN~\citep{wei2018good}, A2RL~\citep{li2018a2} on images from Unsplash~\cite{unsplash2025} for free-form cropping.
  }
  \label{fig:qual}
  \vspace{-0.5em}
\end{figure*}

\noindent\textbf{Number of refinement iteration and VLM model temperature.} Fig.~\ref{fig:number_feedback2} illustrates the effect of the number of refinement iterations $L$ and VLM model temperature on the GAICD~\citep{zeng2020grid} validation dataset for free-form cropping. Since performance peaks after approximately two iterations, we set the number of iterations $L$ to 2. The temperature value controls the randomness of VLM reasoning, with higher temperatures leading to more varied reasoning. We could see temperature 0.05 gives better IoU. So we set temperature to 0.05 for our experiments.

\noindent \textbf{Scorer.}
For evaluating the aesthetics of each crop, we utilize the VILA-R~\citep{ke2023vila} model as our aesthetic scorer. Trained on the AVA~\citep{murray2012ava} dataset, this model specializes in image aesthetic assessment, providing evaluations based on factors such as perspectives, compositions, and color contrast. To measure the content preserving, we calculate the cosine similarity between image CLIP embeddings from cropped image and original image. We also consider the area size of the cropped region as one indicator for content preserving. The area scorer $A = \frac{H_{crop}W_{crop}}{HW} \in [0, 1]$, where $H$, $W$, $H_{crop}$, $W_{crop}$ are the height and width of the input image and cropped image respectively. These scorers are normalized to the range of 0 to 1, and we use different combinations as final score.

\subsection{Comparison with Baselines}
\label{subsec:baseline}

\noindent \textbf{Free-form image cropping.} Tab.~\ref{tab:tab_crop} presents a quantitative comparison between \ours{} and other training-based baselines on the GAICD~\citep{zeng2020grid} dataset. Remarkably, \ours{} outperforms training-based methods by a large margin with only a few in-context learning examples and no training.

Fig.~\ref{fig:qual} shows a visual comparison between \ours{} and other free-form image cropping baselines, namely  GAIC~\citep{zeng2020grid}, VPN~\citep{wei2018good}, A2RL~\citep{li2018a2}. The images are generated using the released codes of these methods. Overall, our approach produces more visually appealing results.

\begin{table}[h!]
    \centering
    \resizebox{0.75\linewidth}{!}{%
      \begin{tabular}{@{}lccc@{}}
        \toprule
        Model  & Training-Free & IoU $\uparrow$ & Disp $\downarrow$ \\
        \midrule
        A2RL~\citep{li2018a2} & \xmark & 0.667 & 0.0887  \\
        VFN~\citep{chen2017learning} & \xmark & 0.669 & 0.0887 \\
        VPN~\citep{wei2018good} & \xmark &0.704  & 0.0699 \\
        VEN~\citep{wei2018good} & \xmark &0.691  & 0.0765   \\
        LVRN~\citep{lu2019listwise} & \xmark & 0.696 & 0.0765 \\
        GAIC~\citep{zeng2020grid} & \xmark & 0.712 & 0.0696 \\
        SAC-Net~\citep{yang2023focusing} &\xmark & 0.767 & 0.0491 \\
        \midrule
        \ours{} (Ours) & \cmark &\bfseries 0.769 & \bfseries 0.0372 \\
        \bottomrule
      \end{tabular}}
      \vspace{-0.5em}
      \caption{Quantitative comparison on the SACD~\cite{yang2023focusing} dataset in subject-aware cropping task.
      }
      \label{tab:sacd}
      \vspace{-1.0em}
\end{table}

\begin{figure*}[t!]
  \centering
  \includegraphics[trim=0 0 0 0,clip,width=\linewidth]{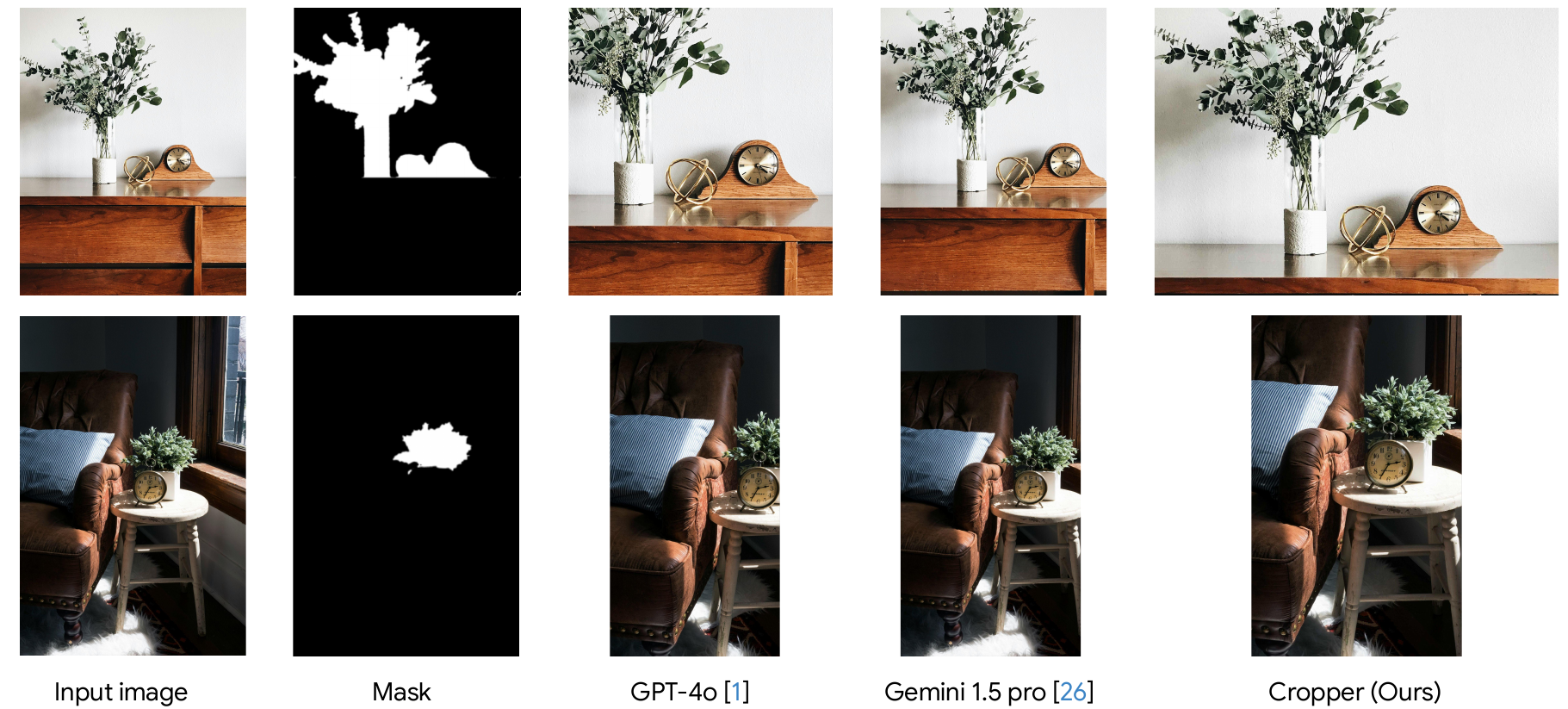}
 \vspace{-1em}
  \caption{Visual comparisons on the subject-aware image cropping. 
  Cropper preserves the important contents better than directly using VLMs, such as GPT-4o~\cite{achiam2023gpt} and Gemini 1.5 Pro~\cite{reid2024gemini}. All input images are from Unsplash~\cite{unsplash2025}.
  }

  \label{fig:sacd}
\end{figure*}

\begin{figure*}[t!]
  \centering
     \includegraphics[width=\linewidth]{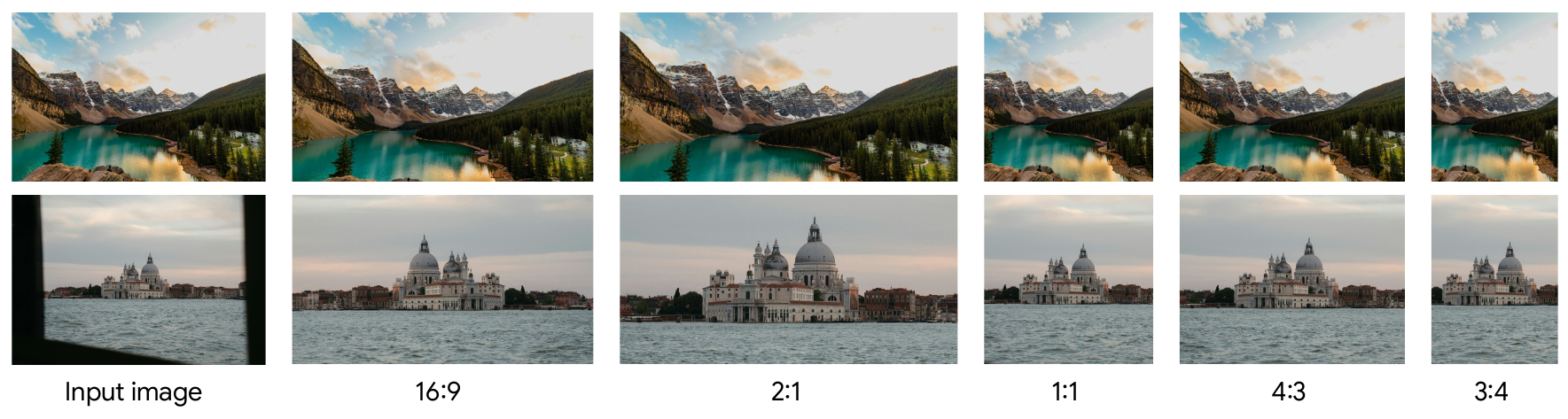}
  \vspace{-0.5em}
  \caption{Example crops from \ours{} for aspect ratio-aware cropping. This shows that our method can generate crops with the desired aspect ratios. All input images are from Unsplash~\cite{unsplash2025}.
}
 \vspace{-0.5em}
  \label{fig:gnmc}
  \vspace{-0.5em}
\end{figure*}

\noindent \textbf{Subject-aware image cropping.} Tab.~\ref{tab:sacd} shows the quantitative comparison on the SACD~\cite{yang2023focusing} dataset, where \ours{} surpasses all other training-based baselines. The reported numbers for other methods are directly taken from the baseline papers. To visually demonstrate the effectiveness of \ours{}, we provide visual samples from \ours{} in Fig.~\ref{fig:sacd}. We show the zero-shot inference results from GPT-4o~\citep{achiam2023gpt} and Gemini 1.5 Pro~\cite{team2023gemini}. GPT-4o is prompted with chain-of-thoughts to crop out the main subject within the image, such as~\textit{``Think step-by-step about finding visually pleasing crops.''}. However, it struggles to generate good crops. For example, the crop from GPT-4o in the first row cuts the main subject, while our cropped image effectively captures the subject of interest.

\noindent \textbf{Aspect ratio-aware image cropping.} Tab.~\ref{tab:gnmc2} shows quantitative comparison results on the FCDB~\citep{9156674} dataset for the aspect ratio-aware cropping task. \ours{} outperforms other baselines in both IoU and Disp, indicating that \ours{} is more adept at cropping the image according to the desired aspect ratio. Fig.~\ref{fig:gnmc} shows example crops from \ours{} for different aspect ratio, illustrating qualitatively that \ours{} can generate crops that possess both good aesthetics and adhere to the specified aspect ratio.

\begin{table}[h!]
      \centering
      \resizebox{0.75\linewidth}{!}{%
      \begin{tabular}{@{}lccc@{}}
        \toprule
        Model  & Training-free & IoU $\uparrow$ & Disp $\downarrow$ \\
        \midrule
        GAIC~\citep{zeng2020grid} & \xmark & 0.673 & 0.064 \\
        A2RL~\citep{li2018a2} & \xmark & 0.695 & 0.073 \\
         VPN~\citep{wei2018good} & \xmark & 0.716 & 0.068\\
        Mars~\citep{9156674} & \xmark & 0.735 & 0.062 \\
        \midrule
        \ours{} (Ours) & \cmark & \bfseries 0.756  & \bfseries 0.053 \\ 
        \bottomrule
      \end{tabular}}
      \caption{Quantitative comparison on the FCDB dataset~\citep{chen2017quantitative} for aspect ratio-aware cropping task. For other methods, we follow~\citep{9156674} to report the modified aspect ratio specified results, which are better than the ones in the original papers. 
      }
      \label{tab:gnmc2}
  \vspace{-1.0em}
\end{table}

\subsection{Ablation study}
\label{subsec:ablation}
\noindent \textbf{Scorer.} Tab.~\ref{tab:ablate_free_from} presents ablation study for different scorers. We experiment with different combinations of VILA~\cite{ke2023vila}, normalized area score, and CLIP~\citep{radford2021learning}. 
Both ``VILA+Area'' and ``VILA+Area+CLIP'' give the best tradeoff for different metrics. We choose ``VILA+Area'' for simplicity.

\begin{table}[h!]
      \centering
      \resizebox{\linewidth}{!}{%
      \begin{tabular}{@{}ccccccccc@{}}
        \toprule
    \multicolumn{3}{l}{Scorer} & \multicolumn{6}{c}{Metrics} \\
    \cmidrule(lr){1-3} \cmidrule(lr){4-9}
    VILA~\cite{ke2023vila} & Area & CLIP~\citep{radford2021learning} & IoU $\uparrow$ & $\overline{Acc}_{5}\uparrow$ & $\overline{Acc}_{10}\uparrow$ & $\overline{SRCC}\uparrow$ & $\overline{PCC}\uparrow$ & Avg$\uparrow$ \\
    \midrule
        \cmark & \xmark & \xmark & 0.748 & 83.6& 95.7&0.901&0.860&0.852\\
        \xmark & \cmark & \xmark & 0.752 & 83.9&96.0&0.882&0.838&0.854 \\
        \xmark & \xmark & \cmark & 0.751 & 81.2 &95.1&0.884&0.833&0.846\\
        \cmark & \cmark & \xmark & 0.748 & \textbf{84.3} & \textbf{96.5}& 0.904&0.860&\textbf{{0.864}} \\
        \cmark & \xmark & \cmark & \textbf{0.754} & 82.3 & 95.8 &0.902&0.850&0.858\\
        \xmark & \cmark & \cmark & 0.752 & 83.9 &96.1&0.902&0.858&0.862 \\
        \cmark & \cmark & \cmark & 0.753 & 83.2 &96.0&\textbf{0.907}&\textbf{0.869}&\textbf{0.864} \\
        \bottomrule
      \end{tabular}}
      \vspace{-0.5em}
      \caption{Ablation study for different scorers on the GAICD~\citep{zeng2020grid} test dataset for free-form cropping.}
      \label{tab:ablate_free_from}
\end{table}

\noindent \textbf{In-context learning.} We compared the results of our method with another two different approaches: 1) zero-shot Gemini-1.5-pro~\cite{reid2024gemini}; 2)~\ours{} with random retrieval for in-context learning examples on the the GAICD test set~\citep{zeng2020grid} for free-form cropping in Tab.~\ref{tab:free_form_max_select}. Our approach achieves the best performance.

\begin{table}[h!]
    \centering
     \resizebox{0.35\textwidth}{!}{
    \begin{tabular}{lcc}
    \toprule
    Method  &  IoU $\uparrow$ & Disp $\downarrow$ \\ 
    \midrule
    Zero-shot Gemini 1.5 Pro~\cite{reid2024gemini}  &  0.509 &  0.1385 \\
    \ours{} with random retrieval  &  0.740 & 0.0660 \\
    \ours{} with CLIP top-$S$ &  \textbf{0.748} & \textbf{0.0635} \\
    \bottomrule
    \end{tabular}}
    \caption{Ablation study for in-context learning. Comparing our methods with zero-shot Gemini 1.5 Pro~\cite{reid2024gemini}, and~\ours{} with random retrieving in-context learning examples on the GAICD test set~\citep{zeng2020grid} for free-form cropping.
    }
    \label{tab:free_form_max_select}
    \vspace{-0.5em}
\end{table}

\noindent \textbf{Iterative refinement.} We show the ablation study on proposed iterative refinement for free-form cropping on both GAICD test set~\citep{zeng2020grid} and FCDB~\citep{9156674} dataset in Tab.~\ref{tab:ablation_ir}. The performance of our model improves significantly with the iterative refinement. 
\begin{table}[h!]
  \centering
    \resizebox{0.8\linewidth}{!}{
    \begin{tabular}{clcc}
        \toprule
        Dataset                & Model                    & IoU $\uparrow$   & Disp $\downarrow$   \\ \midrule
        \multirow{2}{*}{GAICD~\citep{zeng2020grid}} & Cropper w/o Iter Refine. & 0.722 & 0.0679 \\
                               & Cropper (ours)           & \textbf{0.748} & \textbf{0.0635} \\ \midrule
        \multirow{2}{*}{FCDB~\citep{chen2017quantitative}}  & Cropper w/o Iter Refine. & 0.642 & 0.0925 \\
                               & Cropper (ours)           & \textbf{0.667} & \textbf{0.0865} \\ \bottomrule
        \end{tabular}}
   \caption{Ablation study on iterative refinement.
  }
  \label{tab:ablation_ir}
 \vspace{-1.0em}
\end{table}

\noindent \textbf{Final output selection.} We investigate the choice between selecting the output from the final iteration and selecting the output with the highest score across all iterations.
We report the results in Tab.~\ref{tab:free_form_max_select3}. For free-form cropping and subject-aware cropping, we find that using the prediction from the final iteration yields better performance, while aspect ratio-aware is different. 
\begin{table}[h!]
    \centering
    \resizebox{0.45\textwidth}{!}{
    \begin{tabular}{lcccccc}
    \toprule
    & \multicolumn{2}{c}{Free-form} & \multicolumn{2}{c}{Subject} & \multicolumn{2}{c}{Aspect-ratio} \\
    Method  & IoU $\uparrow$ & Disp $\downarrow$ & IoU $\uparrow$ & Disp $\downarrow$ & IoU $\uparrow$ & Disp $\downarrow$\\ 
    \midrule
    Highest score across all iters. &  0.714 & 0.0843 & 0.760 & 0.0381 & \textbf{0.756} & \textbf{0.0529} \\
    From final iter.  &  \textbf{0.748} & \textbf{0.0635} & \textbf{0.769} & \textbf{0.0372} & 0.714 & 0.0632 \\
    \bottomrule
    \end{tabular}}
    \caption{Comparison of selection strategies for free-form cropping on GAICD test set~\citep{zeng2020grid}, subject-aware cropping on the SACD~\cite{yang2023focusing} dataset, and aspect-ratio aware cropping on the FCDB dataset~\citep{chen2017quantitative}.}
    \label{tab:free_form_max_select3}
\end{table}

\noindent\textbf{Vision-language models.} We first evaluate the robustness of various VLMs by performing free-form cropping on the GAICD~\citep{zeng2020grid} dataset using open-source models, such as Mantis-8B Idefics2~\cite{jiang2024mantis}, trained on the Mantis-Instruct~\cite{jiang2024mantis} dataset to perform multi-image tasks. For Mantis-8B-Idefics2~\cite{jiang2024mantis}, we use the same prompt as the free-form cropping task with slightly different parameters: 10 training examples, 5 output crops, and 2 iterations. Additionally, we assess Gemini-1.5-flash~\cite{reid2024gemini}, a lighter variant of Gemini-1.5-pro~\cite{reid2024gemini}. By using Gemini-1.5-flash~\cite{reid2024gemini}, we reduce latency from 5.83 seconds (Gemini-1.5-pro~\cite{reid2024gemini}) to 2.65 seconds. As shown in Tab.~\ref{tab:ablation_vlm}, while lighter VLMs improve efficiency, we observe that larger model yields better cropping performance. Also better VLMs achieve better performance, i.e., results from Gemini is better than Mantis-8B-Idefics2~\cite{jiang2024mantis}.


\begin{table}[h!]
\centering
\resizebox{\linewidth}{!}{%
\begin{tabular}{lccccccc}
\toprule
Model & $\overline{Acc}_{5}$$\uparrow$ & $\overline{Acc}_{10}$$\uparrow$ & $\overline{SRCC}$$\uparrow$ & $\overline{PCC}$$\uparrow$ & IoU$\uparrow$ & Avg$\uparrow$ \\
\midrule
 \ours{} with Mantis-8B-Idefics2~\cite{jiang2024mantis}  & 80.2 & 88.6 & 0.874  & 0.797 & 0.672 & 0.806 \\
 \ours{} with Gemini-1.5-flash~\cite{reid2024gemini} & \textbf{87.2} & \textbf{96.7} & 0.805 & 0.758 & \textbf{0.781} & 0.837 \\
 \ours{} with Gemini-1.5-pro~\cite{reid2024gemini} & 84.3 & 96.5 & \textbf{0.904} & \textbf{0.860} & 0.748 & \textbf{0.864} \\
\bottomrule
\end{tabular}%

}
\vspace{-0.5em}
\caption{Comparing different vision-language models. }
\label{tab:ablation_vlm}
\vspace{-0.8em}

\end{table}

\subsection{User study}
\label{sec:user}
We conduct a user study on 200 test images from the GAICD dataset~\cite{zeng2020grid}. Our methods are compared against A2RL~\citep{li2018a2}, GAIC~\citep{zeng2020grid}, and CGS~\citep{li2020composing}.
For each test image, we show the input, our result and the result from one of three methods, and ask the user to~\textit{``select one image that preserves the most important content from the source image.''}. For each test image, we asked five different users to provide ratings, resulting in a total of $5\times200=1000$ votes.
We show results in Tab.~\ref{tab:user_study}. \ours{} demonstrates a clear superiority over the baseline methods, outperforming them by a significant margin.

\begin{table}[h!]
    \centering
     \resizebox{0.4\textwidth}{!}{\begin{tabular}{lcc}
    \toprule
    Choice  & Baseline ($\%$) & \ours{} ($\%$) \\
    \midrule
    A2RL~\citep{li2018a2} v.s. \ours{}  &  39.2 & \textbf{60.8} \\
    GAIC~\citep{zeng2020grid} v.s. \ours{}  &  37.8 & \textbf{62.2} \\
    CGS~\citep{li2020composing} v.s. \ours{}  &  36.6 & \textbf{63.4} \\
    \bottomrule
    \end{tabular}}
    \vspace{-0.2em}
    \caption{User study on 200 images from the GAICD test set~\cite{zeng2020grid}. We compare with A2RL~\citep{li2018a2}, GAIC~\citep{zeng2020grid} and CGS~\citep{li2020composing}.
    \ours{} shows a superiority over other baselines. }
    \vspace{-5mm}
    \label{tab:user_study}
\end{table}

%% file: sec/6-conclusion.tex
\section{Conclusion \& Limitation}
\vspace{-0.0em}
\label{sec:conclusion}
The paper presents \ours{}, a novel approach to image cropping that leverages in-context learning and vision-language models to achieve superior performance across various cropping tasks. It presents a novel training-free unified approach for tasks like free-form cropping, subject-aware cropping, and aspect ratio-aware cropping. Through extensive experimentation and comparison with existing baselines, \ours{} demonstrates remarkable effectiveness and efficiency, outperforming counterparts with only a few in-context learning examples. Ablation studies show that the proposed visual prompt retrieval strategy and iterative crop refinement approach effectively harness the power of VLMs for effective ICL for cropping. \ours{} has shown effectiveness in improving image cropping performance, but its inference speed constrained by finding suitable in-context examples in a dataset $D$. Advancements of visual retrieval will directly enhance the capabilities of \ours{}.

%% file: supplementary_arxiv.tex
\maketitlesupplementary

\noindent This supplementary material provides:

\begin{itemize}[leftmargin=*]
\item Sec.~\hyperref[sec:a]{A}: We present the implementation details, additional comparison results, additional qualitative results. 
\item Sec.~\hyperref[sec:b]{B}: We present the implementation details, additional ablation study, and additional qualitative results for subject-aware cropping task.
\item Sec.~\hyperref[sec:c]{C}: We present the implementation details, additional ablation study, and additional qualitative results for aspect ratio-aware cropping task.
\item Sec.~\hyperref[sec:d]{D}: We present details about user study.
\end{itemize}

\section*{A. Free-form Cropping}
\label{sec:a}

\subsection*{A.1. Implementation details}
We show the prompt for zero-shot cropping using Gemini 1.5 Pro~\cite{reid2024gemini} in Tab.~\ref{tab:prompt_free_form_gemini}.
\begin{table}[h]
    \centering
    \resizebox{0.4\textwidth}{!}{
    \begin{tabularx}{\linewidth}{lX}
        \toprule
        \textbf{Prompt \& Output} & \textbf{Instruction}  \\
        \midrule
        Initial Prompt & Localize the aesthetic part of the image. $(x_1,y_1,x_2,y_2)$ represents the region. $x_1$ and $x_2$ are the left and right most positions, normalized into 1 to 1000, where 1 is the left and 1000 is the right. $y_1$ and $y_2$ are the top and bottom positions, normalized into 1 to 1000 where 1 is the top and 1000 is the bottom. \\
          & Please propose a new region $(x_1,y_1,x_2,y_2)$ \\ 
          \textbf{Output} & $(\hat{\hat{x}}_1,\hat{\hat{y}}_1,\hat{\hat{x}}_2,\hat{\hat{y}}_2)$ \\
        \bottomrule
    \end{tabularx}}
    \caption{Prompt for zero-shot cropping with Gemini 1.5 Pro~\cite{reid2024gemini}. }
    \label{tab:prompt_free_form_gemini}
\end{table}

Tab.~\ref{tab:fcdb} shows further comparison on the FCDB~\citep{chen2017quantitative} dataset for free-form cropping.

\begin{table}[h!]
    \centering
    \footnotesize
    \begin{tabular}{@{}lcccc@{}}
      \toprule
      Model  & Training-Free & Training Set & IoU $\uparrow$ & Disp $\downarrow$ \\
      \midrule
      A2RL~\citep{li2018a2} & \xmark & AVA & 0.663 & 0.089   \\
      A3RL~\citep{li2019fast}& \xmark & AVA & 0.696 & 0.077   \\
      VPN~\citep{wei2018good} & \xmark& CPC & 0.711  & 0.073 \\
      VEN~\citep{wei2018good} & \xmark& CPC & 0.735  & 0.072   \\
      ASM~\citep{tu2020image}& \xmark & CPC & 0.749  & 0.068 \\
      GAIC~\citep{zeng2020grid} & \xmark& GAICD & 0.672 & 0.084  \\
      CGS~\citep{li2020composing} & \xmark& GAICD & 0.685 & 0.079   \\
      TransView~\citep{pan2021transview} & \xmark& GAICD & 0.682 & 0.080 \\
      Chao et al.~\citep{wang2023image} & \xmark& GAICD &  0.695 & 0.075  \\
      \midrule
       \ours{} (Ours) & \cmark & GAICD &  0.667  & 0.087 \\
      \bottomrule
    \end{tabular}
    \caption{Quantitative comparison among different methods for free-form image cropping on the FCDB~\citep{chen2017quantitative} dataset. 
    \ours{} shows competitive performance as a \emph{training-free} approach.
    }
    \label{tab:fcdb}
\end{table}




    
\subsection*{A.2. Additional qualitative results}
\begin{figure*}[t]
    \centering
    \includegraphics[trim=0 0 0 0,clip,width=0.8\textwidth]{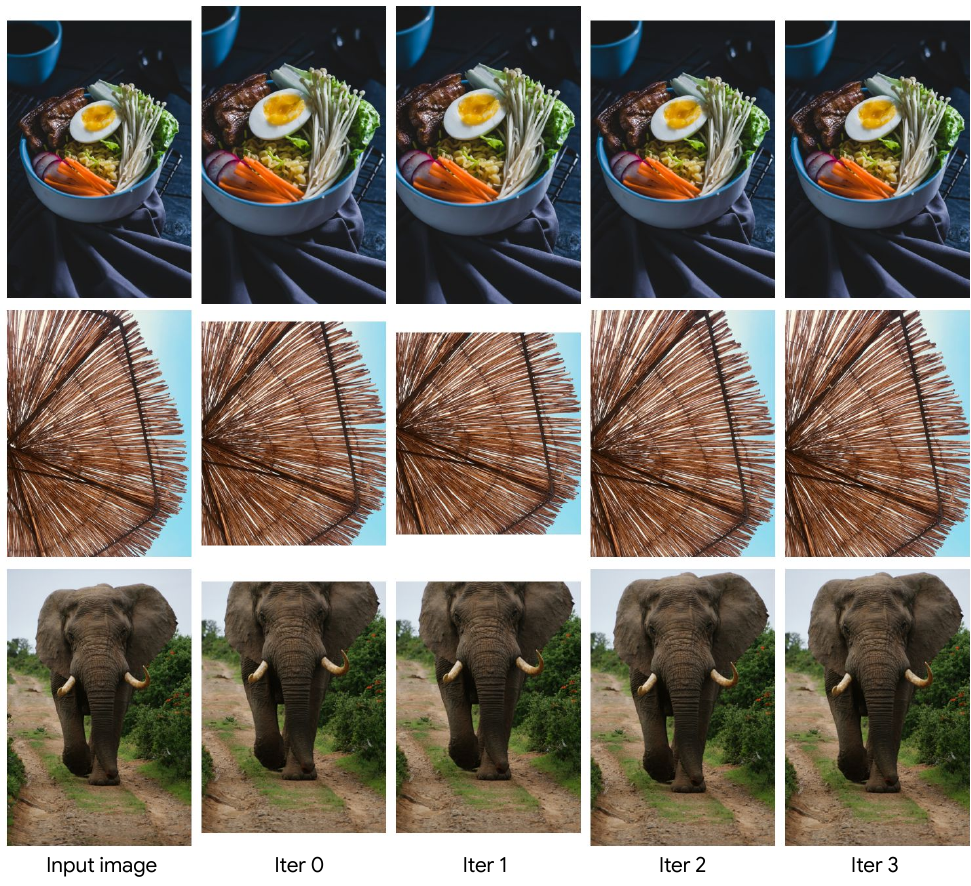}
    \caption{Results from each iteration for free-form cropping using~\ours{}. The iteration process demonstrated progressive convergence, resulting in improved crop quality. All input images are from Unsplash~\cite{unsplash2025}.}
    \label{fig:free-form-iterations}
\end{figure*}

\noindent \textbf{Iterative update.} We showcase some intermediate results of the iterative refinement in Fig.~\ref{fig:free-form-iterations}. 
Our method progressively refines the predicted crops, achieving increasing accuracy and better overlap with the ground-truth cropping box in each iteration.

\begin{figure*}[t]
    \centering
    \includegraphics[width=.8\textwidth]{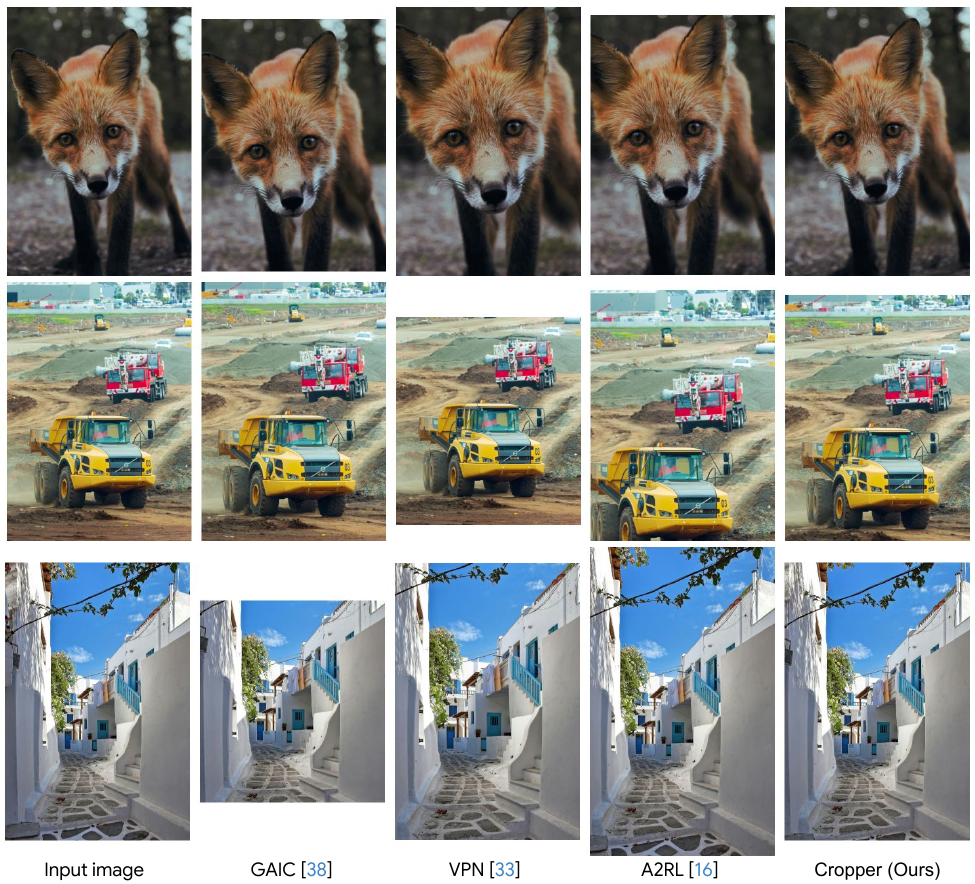}
    \caption{Comparing with GAIC~\citep{zeng2020grid}, VPN~\citep{wei2018good} and A2RL~\citep{li2018a2} for free-form cropping on images from Unsplash~\cite{unsplash2025}.
    }
    \label{fig:supp_freeform_examples}
\end{figure*}

\noindent \textbf{Qualitative comparison.} We present additional results in Fig.~\ref{fig:supp_freeform_examples}. Our method generates better visual pleasing crops. 

\section*{B. Subject-aware Cropping}
\label{sec:b}
\subsection*{B.1. Prompts}
We show the details of the prompts for the subject-aware cropping in Tab.~\ref{tab:prompt_subject_aware}. The goal is to get accurate coordinates of the crop $(\hat{\hat{x}}_1,\hat{\hat{y}}_1,\hat{\hat{x}}_2,\hat{\hat{y}}_2)$. In the initial prompt, we use 30 in-context~(ICL) examples for image cropping for 10 iterations. 10 examples are ranked by the scorer and we use top-5 crops for our task, the format of image $i$'s $j$-th crop is defined as $(x_1^{i, j}, y_1^{i, j}, x_2^{i, j}, y_2^{i, j})$. Intermediate results of initial prompt are coordinates of 5 crops. Subsequently, the crop is iteratively refined by accumulating the context into prompts, using refinement prompt. Note that scorer is ``VILA+Area''.

\begin{table}[h!]
    \centering
    \scriptsize
    \begin{tabularx}{\linewidth}{lX}
        \toprule
        \textbf{Prompt \& Output} & \textbf{Instruction}  \\
        \midrule
        Initial Prompt & Find visually appealing crop. Each region is represented by $(x_1,y_1,x_2,y_2)$ coordinates. $x_1, x_2$ are the left and right most positions, normalized into 0 to 1, where 0 is the left and 1 is the right. $y_1, y_2$ are the top and bottom positions, normalized into 0 to 1 where 0 is the top and 1 is the bottom. \\
        & \{image 1\} $((c^1_{x}, c^1_{y}),x^{1}_1,y^{1}_1,x^{1}_2,y^{1}_2)$, \\
        & \{image 2\}, $((c^2_{x}, c^2_{y}),x_1^{2},y_1^{2},x_2^{2},y_2^{2})$, \\
        & ... \\ 
        & \{image $S$\}, $((c^{S}_{x}, c^{S}_{y}),x_1^{S},y_1^{S},x_2^{S},y_2^{S})$, \\
        & \{Query image\}, $(c_x, c_y)$ \\
        \textbf{Output} & $(\hat{x}_1,\hat{y}_1,\hat{x}_2,\hat{y}_2)$  \\
         \midrule
        Iterative Crop  &  Localize aesthetic part of image. The region is \\
        Refinement Prompt &  represented by ($x_1$,$y_1$,$x_2$,$y_2$). $x_1$, $x_2$ are the left and right most positions, normalized into 0 to 1, where 0 is the left and 1 is the right. $y_1$, $y_2$ are the top and bottom positions, normalized into 0 to 1 where 0 is the top and 1 is the bottom. We provide several images here. \\ 
        & \{Cropped image 1\} Output:  $(\hat{x}_1^1,\hat{y}_1^1,\hat{x}_2^1,\hat{y}_2^1)$\\
        & \{Cropped image 2\} Output:  $(\hat{x}_1^2,\hat{y}_1^2,\hat{x}_2^2,\hat{y}_2^2)$ \\
        & ... \\
        & \{Cropped image $R$\} Output: $(\hat{x}_1^R,\hat{y}_1^R,\hat{x}_2^R,\hat{y}_2^R)$ \\
          & Propose different crop. The region should be represented by ($x_1$,$y_1$,$x_2$,$y_2$). Output:\\ 
          \textbf{Output} & $(\hat{\hat{x}}_1,\hat{\hat{y}}_1,\hat{\hat{x}}_2,\hat{\hat{y}}_2)$ \\
        \bottomrule
    \end{tabularx}
    \caption{VLM prompt used for subject-aware cropping.
    }
    \vspace{-5mm}
    \label{tab:prompt_subject_aware} 
\end{table}

\subsection*{B.2. Ablation study of scores}
We show the ablation study of scorer on the subject-aware cropping in Tab.~\ref{tab:ablate_subject_aware}.
With ``VILA+Area'', our proposed method achieves the best performance.

\begin{table}[h!]
      \centering
      \resizebox{0.7\linewidth}{!}{%
      \begin{tabular}{@{}ccccc@{}}
        \toprule
        VILA~\cite{ke2023vila}  & Area & CLIP~\citep{radford2021learning} & IoU $\uparrow$ & Disp $\downarrow$ \\
        \midrule
        \cmark & \xmark & \xmark & 0.753 & 0.0413 \\
        \xmark & \cmark & \xmark & 0.755 & 0.0402 \\
        \xmark & \xmark & \cmark & 0.749 & 0.0417 \\
        \cmark & \cmark & \xmark & \textbf{0.769} & \textbf{0.0372} \\
        \cmark & \xmark & \cmark & 0.751 & 0.0401 \\
        \xmark & \cmark & \cmark & 0.754 & 0.0394 \\
        \cmark & \cmark & \cmark & 0.766 & 0.0379 \\
        \bottomrule
      \end{tabular}}
      \caption{Ablation study for scores on the subject-aware cropping.  
      \ours{} achieves the best performance with VILA~\cite{ke2023vila} + Area score.
      }
      \label{tab:ablate_subject_aware}
\end{table}

\subsection*{B.3. Additional qualitative results}
\begin{figure*}[t]
    \centering
    \includegraphics[trim=0 0 0 0,clip,width=1\textwidth]{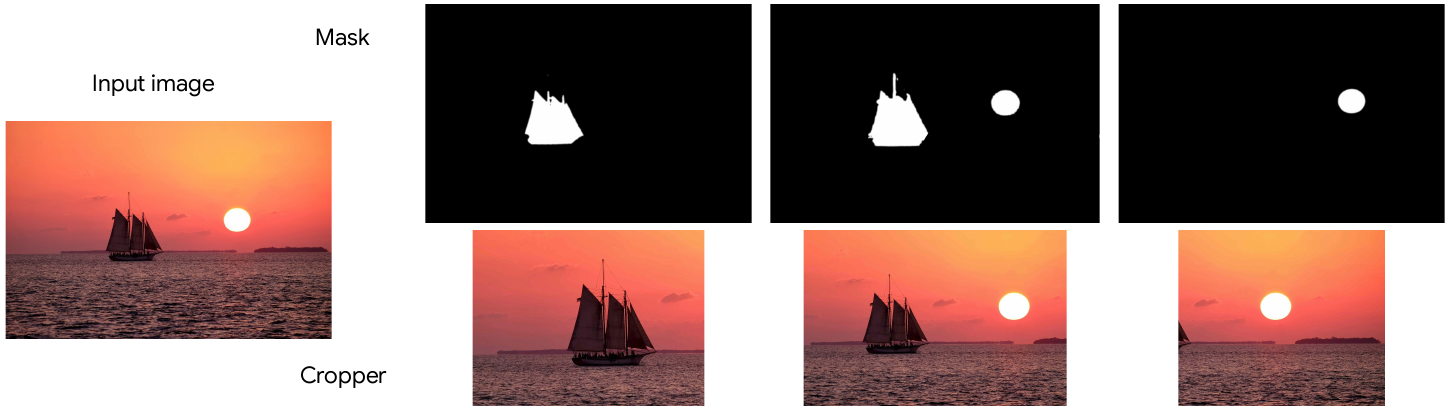}
    \caption{Qualitative results of subject-aware cropping. The result shows that our method can generate crops on different subjects. The input image is from Unsplash~\cite{unsplash2025}.}
    \label{fig:supp_subject_aware}
\end{figure*}

We showcase more results in Fig.~\ref{fig:supp_subject_aware}. Our method demonstrates subject awareness, enabling the generation of high-quality cropped images.

\section*{C. Aspect-ratio aware Cropping}
\label{sec:c}
\subsection*{C.1. Prompts}
We show the details of the prompts for the aspect ratio-aware cropping in Tab.~\ref{tab:mul2}. For this task, we use the following hyperparameter: number of in-context learning examples$S = 10$, number of crops $R = 6$, number of iteration $L=2$, temperature = 0.05.

\begin{table}[h!]
    \centering
    \scriptsize
    \begin{tabularx}{\linewidth}{lX}
        \toprule
        \textbf{Prompt \& Output} & \textbf{Instruction}  \\
        \midrule
        Initial Prompt & Find visually appealing crop.  Give the best crop in the form of a crop box and make sure the crop has certain width:height. Box is a 4-tuple defining the left, upper, right, and lower pixel coordinate in the form of $(x_1,y_1,x_2,y_2)$. Here are some example images, its size, and crop w:h triplets and their corresponding crops. \\
        & \{image 1\}, size $(w_1, h_1)$, crop ratio ($r_1$), output $(x^{1}_1,y^{1}_1,x^{1}_2,y^{1}_2)$,  \\
        & \{ \text{image 2} \}, \text{size } $(w_2, h_2)$, \text{crop ratio } ($r_2$), \text{output } $(x^{2}_1,y^{2}_1,x^{2}_2,y^{2}_2)$, \\
        & ... \\ 
        & \{image $S$\},  \text{size } $(w_{S}, h_{S})$, \text{crop ratio } ($r_{S}$), \text{output } $(x^{S}_1,y^{S}_1,x^{S}_2,y^{S}_2)$, \\&
\{Now Give the best crop in the form of a crop box for the following image. Give $R$ possible best crops.\}\\
        & \{Query image\}, size $(w, h)$, crop ratio ($r$) \\
        \textbf{Output} & $(\hat{x}_1^1,\hat{y}_1^1,\hat{x}_2^1,\hat{y}_2^1)$,$(\hat{x}_1^2,\hat{y}_1^2,\hat{x}_2^2,\hat{y}_2^2)$, ..., $(\hat{x}_1^T,\hat{y}_1^T,\hat{x}_2^T,\hat{y}_2^T)$  \\
         \midrule
        Iterative Crop  &  Initial Prompt + Example Image: \{Query image\}; \\
        Refinement Prompt & Crop ratio: $r$; Example output:  \\
        &\{Cropped image 1\} $(\hat{x}_1^1,\hat{y}_1^1,\hat{x}_2^1,\hat{y}_2^1)$, \\
        & \{Cropped image 2\} $(\hat{x}_1^2,\hat{y}_1^2,\hat{x}_2^2,\hat{y}_2^2)$ \\
        & ... \\
        & \{Cropped image $R$\} $(\hat{x}_1^R,\hat{y}_1^R,\hat{x}_2^R,\hat{y}_2^R)$ \\
        & Propose a different better crop with the given ratio. Output:\\ 
          \textbf{Output} & $(\hat{\hat{x}}_1,\hat{\hat{y}}_1,\hat{\hat{x}}_2,\hat{\hat{y}}_2)$ \\
        \bottomrule
    \end{tabularx}
    \caption{VLM prompt used for aspect ratio-aware cropping.} 
    \label{tab:mul2}  
\end{table}

\subsection*{C.2. Ablation study of scores}
We show the ablation study of scores on the aspect ratio-aware cropping in Tab.~\ref{tab:gnmc}.
With CLIP score only, our proposed method achieves the best performance.

\begin{table}[h!]
    \centering
    \resizebox{0.7\linewidth}{!}{%
    \begin{tabular}{@{}cccccc@{}}
    \toprule
    VILA~\cite{ke2023vila} & Area & CLIP~\cite{radford2021learning} & IoU $\uparrow$ & Disp $\downarrow$ \\
    \midrule
    \cmark & \xmark & \xmark & 0.718 & 0.0631 \\
    \xmark & \cmark & \xmark & 0.713 & 0.0630 \\
    \xmark & \xmark & \cmark & \textbf{0.756} & \textbf{0.0529} \\
    \cmark & \cmark & \xmark & 0.716 & 0.0630 \\
    \xmark & \cmark & \cmark & 0.741 & 0.0562 \\
    \cmark & \xmark & \cmark & 0.742 & 0.0560 \\
    \cmark & \cmark & \cmark & 0.729 & 0.0588 \\
    \bottomrule
    \end{tabular}}
    \caption{Ablation study for aspect ratio-aware cropping task. Comparison of combinations of VILA~\cite{ke2023vila}, Area, and CLIP~\cite{radford2021learning} components shows that the CLIP-only configuration achieves the best IoU and Disp values.}
    \label{tab:gnmc}
\end{table}

\subsection*{D. User study}
\label{sec:d}
We include the instructions for users as follows:
\begin{itemize}
    \item Your Task: Carefully analyze the source image and the two output images and SELECT one output.

    \item Content: This refers to the key elements and objects in the image, such as people, buildings, or other recognizable features. The output should keep the important details of these objects as close to the original as possible.

    \item Aesthetics: The output has a sense of aesthetics. It follows common human natures with proper layout. Select the output that not only preserves the content best and fits the aesthetics.

    \item Your Goal: Select the image that, overall, looks the most natural and visually appealing to the source image.
\end{itemize}

%% file: main.bbl
\begin{thebibliography}{45}
\providecommand{\natexlab}[1]{#1}
\providecommand{\url}[1]{\texttt{#1}}
\expandafter\ifx\csname urlstyle\endcsname\relax
  \providecommand{\doi}[1]{doi: #1}\else
  \providecommand{\doi}{doi: \begingroup \urlstyle{rm}\Url}\fi

\bibitem[Achiam et~al.(2023)Achiam, Adler, Agarwal, Ahmad, Akkaya, Aleman, Almeida, Altenschmidt, Altman, Anadkat, et~al.]{achiam2023gpt}
Josh Achiam, Steven Adler, Sandhini Agarwal, Lama Ahmad, Ilge Akkaya, Florencia~Leoni Aleman, Diogo Almeida, Janko Altenschmidt, Sam Altman, Shyamal Anadkat, et~al.
\newblock Gpt-4 technical report.
\newblock \emph{arXiv preprint arXiv:2303.08774}, 2023.

\bibitem[Bar et~al.(2022)Bar, Gandelsman, Darrell, Globerson, and Efros]{bar2022visual}
Amir Bar, Yossi Gandelsman, Trevor Darrell, Amir Globerson, and Alexei Efros.
\newblock Visual prompting via image inpainting.
\newblock In \emph{NeurIPS}, 2022.

\bibitem[Chen et~al.(2016)Chen, Bai, Liang, and Li]{chen2016automatic}
Jiansheng Chen, Gaocheng Bai, Shaoheng Liang, and Zhengqin Li.
\newblock Automatic image cropping: A computational complexity study.
\newblock In \emph{CVPR}, 2016.

\bibitem[Chen et~al.(2017{\natexlab{a}})Chen, Huang, Chang, Tsai, Chen, and Chen]{chen2017quantitative}
Yi-Ling Chen, Tzu-Wei Huang, Kai-Han Chang, Yu-Chen Tsai, Hwann-Tzong Chen, and Bing-Yu Chen.
\newblock Quantitative analysis of automatic image cropping algorithms: A dataset and comparative study.
\newblock In \emph{WACV}, 2017{\natexlab{a}}.

\bibitem[Chen et~al.(2017{\natexlab{b}})Chen, Klopp, Sun, Chien, and Ma]{chen2017learning}
Yi-Ling Chen, Jan Klopp, Min Sun, Shao-Yi Chien, and Kwan-Liu Ma.
\newblock Learning to compose with professional photographs on the web.
\newblock In \emph{MM}, 2017{\natexlab{b}}.

\bibitem[Cheng et~al.(2010)Cheng, Ni, Yan, and Tian]{cheng2010learning}
Bin Cheng, Bingbing Ni, Shuicheng Yan, and Qi Tian.
\newblock Learning to photograph.
\newblock In \emph{MM}, 2010.

\bibitem[Cheng et~al.(2022)Cheng, Lin, and Allebach]{cheng2022re}
Yang Cheng, Qian Lin, and Jan~P Allebach.
\newblock Re-compose the image by evaluating the crop on more than just a score.
\newblock In \emph{WACV}, 2022.

\bibitem[Deng et~al.(2018)Deng, Loy, and Tang]{deng2018aesthetic}
Yubin Deng, Chen~Change Loy, and Xiaoou Tang.
\newblock Aesthetic-driven image enhancement by adversarial learning.
\newblock In \emph{MM}, 2018.

\bibitem[{Gemini Team Google}(2023)]{team2023gemini}
{Gemini Team Google}.
\newblock Gemini: A family of highly capable multimodal models.
\newblock \emph{arXiv preprint arXiv:2312.11805}, 2023.

\bibitem[Guo et~al.(2018)Guo, Wang, Shen, Yan, and Liao]{guo2018automatic}
Guanjun Guo, Hanzi Wang, Chunhua Shen, Yan Yan, and Hong-Yuan~Mark Liao.
\newblock Automatic image cropping for visual aesthetic enhancement using deep neural networks and cascaded regression.
\newblock \emph{IEEE Trans. Multimedia}, 20\penalty0 (8), 2018.

\bibitem[Hong et~al.(2021)Hong, Du, Xian, Lu, Cao, and Zhong]{hong2021composing}
Chaoyi Hong, Shuaiyuan Du, Ke Xian, Hao Lu, Zhiguo Cao, and Weicai Zhong.
\newblock Composing photos like a photographer.
\newblock In \emph{CVPR}, 2021.

\bibitem[Hong et~al.(2024)Hong, Yuan, Gharbi, Fisher, and Fatahalian]{hong2023learning}
James Hong, Lu Yuan, Micha{\"e}l Gharbi, Matthew Fisher, and Kayvon Fatahalian.
\newblock Learning subject-aware cropping by outpainting professional photos.
\newblock In \emph{AAAI}, 2024.

\bibitem[Jiang et~al.(2024)Jiang, He, Zeng, Wei, Ku, Liu, and Chen]{jiang2024mantis}
Dongfu Jiang, Xuan He, Huaye Zeng, Cong Wei, Max Ku, Qian Liu, and Wenhu Chen.
\newblock Mantis: Interleaved multi-image instruction tuning.
\newblock \emph{TMLR}, 2024.

\bibitem[Kao et~al.(2017)Kao, He, and Huang]{kao2017automatic}
Yueying Kao, Ran He, and Kaiqi Huang.
\newblock Automatic image cropping with aesthetic map and gradient energy map.
\newblock In \emph{ICASSP}, 2017.

\bibitem[Ke et~al.(2023)Ke, Ye, Yu, Wu, Milanfar, and Yang]{ke2023vila}
Junjie Ke, Keren Ye, Jiahui Yu, Yonghui Wu, Peyman Milanfar, and Feng Yang.
\newblock Vila: Learning image aesthetics from user comments with vision-language pretraining.
\newblock In \emph{CVPR}, 2023.

\bibitem[Li et~al.(2018)Li, Wu, Zhang, and Huang]{li2018a2}
Debang Li, Huikai Wu, Junge Zhang, and Kaiqi Huang.
\newblock A2-rl: Aesthetics aware reinforcement learning for image cropping.
\newblock In \emph{CVPR}, 2018.

\bibitem[Li et~al.(2019)Li, Wu, Zhang, and Huang]{li2019fast}
Debang Li, Huikai Wu, Junge Zhang, and Kaiqi Huang.
\newblock Fast a3rl: Aesthetics-aware adversarial reinforcement learning for image cropping.
\newblock \emph{TIP}, 28\penalty0 (10), 2019.

\bibitem[Li et~al.(2020{\natexlab{a}})Li, Zhang, and Huang]{9156674}
Debang Li, Junge Zhang, and Kaiqi Huang.
\newblock Learning to learn cropping models for different aspect ratio requirements.
\newblock In \emph{CVPR}, 2020{\natexlab{a}}.

\bibitem[Li et~al.(2020{\natexlab{b}})Li, Zhang, Huang, and Yang]{li2020composing}
Debang Li, Junge Zhang, Kaiqi Huang, and Ming-Hsuan Yang.
\newblock Composing good shots by exploiting mutual relations.
\newblock In \emph{CVPR}, 2020{\natexlab{b}}.

\bibitem[Lu et~al.(2019)Lu, Xing, Cai, and Xu]{lu2019listwise}
Weirui Lu, Xiaofen Xing, Bolun Cai, and Xiangmin Xu.
\newblock Listwise view ranking for image cropping.
\newblock \emph{IEEE Access}, 7, 2019.

\bibitem[Lu et~al.(2022)Lu, Bartolo, Moore, Riedel, and Stenetorp]{lu2021fantastically}
Yao Lu, Max Bartolo, Alastair Moore, Sebastian Riedel, and Pontus Stenetorp.
\newblock Fantastically ordered prompts and where to find them: Overcoming few-shot prompt order sensitivity.
\newblock In \emph{ACL}, 2022.

\bibitem[Mai et~al.(2016)Mai, Jin, and Liu]{mai2016composition}
Long Mai, Hailin Jin, and Feng Liu.
\newblock Composition-preserving deep photo aesthetics assessment.
\newblock In \emph{CVPR}, 2016.

\bibitem[Murray et~al.(2012)Murray, Marchesotti, and Perronnin]{murray2012ava}
Naila Murray, Luca Marchesotti, and Florent Perronnin.
\newblock Ava: A large-scale database for aesthetic visual analysis.
\newblock In \emph{CVPR}, 2012.

\bibitem[Pan et~al.(2021)Pan, Cao, Wang, Lu, and Zhong]{pan2021transview}
Zhiyu Pan, Zhiguo Cao, Kewei Wang, Hao Lu, and Weicai Zhong.
\newblock Transview: Inside, outside, and across the cropping view boundaries.
\newblock In \emph{ICCV}, 2021.

\bibitem[Radford et~al.(2021)Radford, Kim, Hallacy, Ramesh, Goh, Agarwal, Sastry, Askell, Mishkin, Clark, et~al.]{radford2021learning}
Alec Radford, Jong~Wook Kim, Chris Hallacy, Aditya Ramesh, Gabriel Goh, Sandhini Agarwal, Girish Sastry, Amanda Askell, Pamela Mishkin, Jack Clark, et~al.
\newblock Learning transferable visual models from natural language supervision.
\newblock In \emph{ICML}, 2021.

\bibitem[Reid et~al.(2024)Reid, Savinov, Teplyashin, Lepikhin, Lillicrap, Alayrac, Soricut, Lazaridou, Firat, Schrittwieser, et~al.]{reid2024gemini}
Machel Reid, Nikolay Savinov, Denis Teplyashin, Dmitry Lepikhin, Timothy Lillicrap, Jean-baptiste Alayrac, Radu Soricut, Angeliki Lazaridou, Orhan Firat, Julian Schrittwieser, et~al.
\newblock Gemini 1.5: Unlocking multimodal understanding across millions of tokens of context.
\newblock \emph{arXiv preprint arXiv:2403.05530}, 2024.

\bibitem[Rubin et~al.(2022)Rubin, Herzig, and Berant]{rubin2021learning}
Ohad Rubin, Jonathan Herzig, and Jonathan Berant.
\newblock Learning to retrieve prompts for in-context learning.
\newblock In \emph{ACL}, 2022.

\bibitem[Sun and Ling(2013)]{sun2013scale}
Jin Sun and Haibin Ling.
\newblock Scale and object aware image thumbnailing.
\newblock \emph{IJCV}, 104, 2013.

\bibitem[Tu et~al.(2020)Tu, Niu, Zhao, Cheng, and Zhang]{tu2020image}
Yi Tu, Li Niu, Weijie Zhao, Dawei Cheng, and Liqing Zhang.
\newblock Image cropping with composition and saliency aware aesthetic score map.
\newblock In \emph{AAAI}, 2020.

\bibitem[{Unsplash Website}()]{unsplash2025}
{Unsplash Website}.
\newblock Unsplash.
\newblock Accessed: March 21, 2025, URL: https://unsplash.com/.

\bibitem[Wang et~al.(2023{\natexlab{a}})Wang, Niu, Zhang, and Zhang]{wang2023image}
Chao Wang, Li Niu, Bo Zhang, and Liqing Zhang.
\newblock Image cropping with spatial-aware feature and rank consistency.
\newblock In \emph{CVPR}, 2023{\natexlab{a}}.

\bibitem[Wang et~al.(2023{\natexlab{b}})Wang, Wang, Cao, Shen, and Huang]{wang2023images}
Xinlong Wang, Wen Wang, Yue Cao, Chunhua Shen, and Tiejun Huang.
\newblock Images speak in images: A generalist painter for in-context visual learning.
\newblock In \emph{CVPR}, 2023{\natexlab{b}}.

\bibitem[Wei et~al.(2018)Wei, Zhang, Shen, Lin, Mech, Hoai, and Samaras]{wei2018good}
Zijun Wei, Jianming Zhang, Xiaohui Shen, Zhe Lin, Radomir Mech, Minh Hoai, and Dimitris Samaras.
\newblock Good view hunting: Learning photo composition from dense view pairs.
\newblock In \emph{CVPR}, 2018.

\bibitem[Yan et~al.(2013)Yan, Lin, Bing~Kang, and Tang]{yan2013learning}
Jianzhou Yan, Stephen Lin, Sing Bing~Kang, and Xiaoou Tang.
\newblock Learning the change for automatic image cropping.
\newblock In \emph{CVPR}, 2013.

\bibitem[Yang et~al.(2024)Yang, Wang, Lu, Liu, Le, Zhou, and Chen]{yang2024large}
Chengrun Yang, Xuezhi Wang, Yifeng Lu, Hanxiao Liu, Quoc~V Le, Denny Zhou, and Xinyun Chen.
\newblock Large language models as optimizers.
\newblock In \emph{ICLR}, 2024.

\bibitem[Yang et~al.(2023)Yang, Zhou, Cai, Zhang, and Zhang]{yang2023focusing}
Guo-Ye Yang, Wen-Yang Zhou, Yun Cai, Song-Hai Zhang, and Fang-Lue Zhang.
\newblock Focusing on your subject: Deep subject-aware image composition recommendation networks.
\newblock \emph{CVM}, 9\penalty0 (1), 2023.

\bibitem[Zeng et~al.(2019)Zeng, Li, Cao, and Zhang]{zeng2019reliable}
Hui Zeng, Lida Li, Zisheng Cao, and Lei Zhang.
\newblock Reliable and efficient image cropping: A grid anchor based approach.
\newblock In \emph{CVPR}, 2019.

\bibitem[Zeng et~al.(2020)Zeng, Li, Cao, and Zhang]{zeng2020grid}
Hui Zeng, Lida Li, Zisheng Cao, and Lei Zhang.
\newblock Grid anchor based image cropping: A new benchmark and an efficient model.
\newblock \emph{TPAMI}, 44\penalty0 (3), 2020.

\bibitem[Zhang et~al.(2023{\natexlab{a}})Zhang, Haddow, and Birch]{zhang2023prompting}
Biao Zhang, Barry Haddow, and Alexandra Birch.
\newblock Prompting large language model for machine translation: A case study.
\newblock In \emph{ICML}, 2023{\natexlab{a}}.

\bibitem[Zhang et~al.(2024)Zhang, Wang, Li, Nakashima, and Nagahara]{zhang2024instruct}
Jiahao Zhang, Bowen Wang, Liangzhi Li, Yuta Nakashima, and Hajime Nagahara.
\newblock Instruct me more! random prompting for visual in-context learning.
\newblock In \emph{WACV}, 2024.

\bibitem[Zhang et~al.(2013)Zhang, Song, Yang, Zhao, Zhao, and Sebe]{zhang2013weakly}
Luming Zhang, Mingli Song, Yi Yang, Qi Zhao, Chen Zhao, and Nicu Sebe.
\newblock Weakly supervised photo cropping.
\newblock \emph{TMM}, 16\penalty0 (1), 2013.

\bibitem[Zhang et~al.(2005)Zhang, Zhang, Sun, Feng, and Ma]{zhang2005auto}
Mingju Zhang, Lei Zhang, Yanfeng Sun, Lin Feng, and Weiying Ma.
\newblock Auto cropping for digital photographs.
\newblock In \emph{ICME}, 2005.

\bibitem[Zhang et~al.(2022)Zhang, Feng, and Tan]{zhang-etal-2022-active}
Yiming Zhang, Shi Feng, and Chenhao Tan.
\newblock Active example selection for in-context learning.
\newblock In \emph{EMNLP}, 2022.

\bibitem[Zhang et~al.(2023{\natexlab{b}})Zhang, Zhou, and Liu]{zhang2024makes}
Yuanhan Zhang, Kaiyang Zhou, and Ziwei Liu.
\newblock What makes good examples for visual in-context learning?
\newblock In \emph{NeurIPS}, 2023{\natexlab{b}}.

\bibitem[Zhong et~al.(2021)Zhong, Li, Huang, Zhang, Lu, and Wang]{zhong2021aesthetic}
Lei Zhong, Feng-Heng Li, Hao-Zhi Huang, Yong Zhang, Shao-Ping Lu, and Jue Wang.
\newblock Aesthetic-guided outward image cropping.
\newblock \emph{TOG}, 40\penalty0 (6), 2021.

\end{thebibliography}
